\pdfoutput=1

\documentclass[11pt]{article}

\usepackage[final]{acl}

\usepackage{times}
\usepackage{latexsym}

\usepackage[T1]{fontenc}

\usepackage[utf8]{inputenc}
\usepackage{booktabs}
\usepackage{multirow}
\usepackage{graphicx}
\usepackage[table]{xcolor} 

\usepackage{microtype}
\usepackage{booktabs}
\usepackage{kotex}
\usepackage{bbm}
\usepackage{algorithm}
\usepackage{url}
\usepackage{soul}
\usepackage{verbatim}
\usepackage{enumitem}
\usepackage{algorithm}
\usepackage{algorithmic}
\usepackage{amsmath, amssymb}       
\usepackage{mathtools}              
\usepackage{bm} 
\usepackage{subcaption}
\usepackage{multirow}
\usepackage{lipsum}
\usepackage{booktabs}
\usepackage{multirow}
\usepackage{booktabs}
\usepackage{inconsolata}
\usepackage[table]{xcolor}
\usepackage{graphicx}
\usepackage{caption}
\usepackage{tabularx}   
\usepackage{makecell}   
\usepackage{booktabs}   
\usepackage{booktabs,tabularx,siunitx}
\usepackage[x11names]{xcolor} 
\usepackage{booktabs}
\usepackage{multirow}
\usepackage{graphicx}
\usepackage{colortbl}
\usepackage{xcolor}
\usepackage{booktabs}
\usepackage{graphicx}
\usepackage{multirow}
\usepackage{soul}
\usepackage{microtype}

\newcommand{\km}[1]{\textcolor{black}{#1}}
\newcommand{\hj}[1]{\textcolor{black}{#1}}
\newcommand{\nj}[1]{\textcolor{black}{#1}}

\definecolor{bestc}{HTML}{FFD700} 
\definecolor{secondc}{HTML}{E6E6E6} 

\definecolor{mygreen}{HTML}{E29578}
\title{Knowledge Beyond Language:\\Bridging the Gap in Multilingual Machine Unlearning Evaluation}

\author{
  Kyomin Hwang$^{1*}$ \quad
  Hyeonjin Kim$^{1*}$ \quad
  Sangyeon Cho$^{3,4}$ \quad
  Nojun Kwak$^{1,2\dagger}$ \\[0.4em]
  $^1$GSCST, Seoul National University \\
  $^2$AIIS, Seoul National University \\
  $^3$Department of Artificial Intelligence, Chung-Ang University \\
  $^4$Korean Surgical Researcher Foundation, Republic of Korea \\[0.4em]
  \texttt{\{kyomin98, peaceful1, nojunk\}@snu.ac.kr} \quad
  \texttt{whtkddus98@cau.ac.kr}
}

\begin{document}
\maketitle
\renewcommand{\thefootnote}{\fnsymbol{footnote}}
\footnotetext[1]{Equal contribution.}
\footnotetext[2]{Corresponding author.}
\begin{abstract}
\km{\hj{While LLMs are increasingly used} in commercial services, they pose privacy risks such as leakage of sensitive personal\nj{ly identifiable} information (PII). For LLMs trained on multilingual corpora, Multilingual Machine Unlearning (MMU) aims to remove \hj{information} across \hj{multiple} languages. However, prior MMU evaluations fail to capture \hj{such cross-linguistic distribution of information, being largely limited to direct extensions of per-language evaluation protocols.} To this end, we propose two metrics to evaluate the \hj{information} spread across languages: the Knowledge Separability Score (KSS) and the Knowledge Persistence Score (KPS). \hj{KSS measures the overall unlearning quality across multiple languages, while KPS more specifically aims to assess consistent removal of information }\nj{among} different language pairs. We evaluated various unlearning methods in the multilingual setting with these metrics and conducted comprehensive analyses. Through our investigation, we provide insights into unique phenomena exclusive to MMU and offer a new perspective on MMU evaluation.}
\end{abstract}
\section{Introduction}

\begin{figure}[t]
    \centering
    \includegraphics[width=0.9\linewidth]{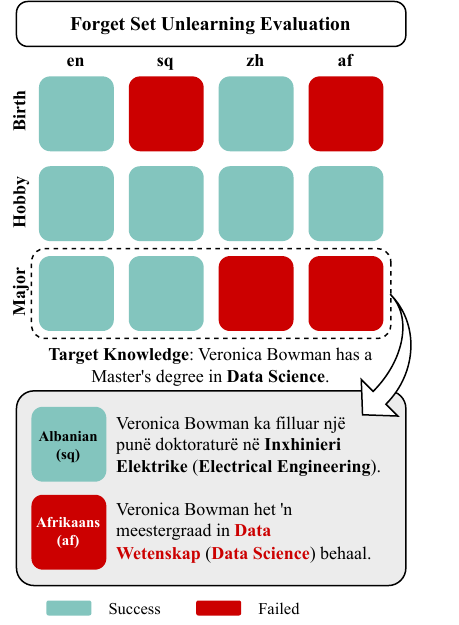}
    \caption{\km{Illustration of the evaluation method in conventional MMU. Existing approaches evaluate knowledge (e.g., Birth, Hobby) independently for each language. Consequently, this language-wise assessment fails to verify whether knowledge \hj{has spread} across different languages has been \nj{successfully removed}.}}
    \label{fig:teaser}
\end{figure}

Machine Unlearning (MU) aims to remove sensitive information from a Large Language Model (LLM)~\cite{mu-survey}. Since \citeauthor{ga} demonstrated the feasibility of unlearning via gradient ascent, subsequent methods have been developed and evaluated on English datasets, focusing on erasing the specified content without degrading overall performance~\cite{npo, grad_diff, tofu, muse}. However, previous works simulate MU with an English-only dataset, leaving a gap to real-world deployment.

To bridge this gap, recent studies have begun to investigate Multilingual MU (MMU)~\cite{lingtea, lau, hwang2025uncovering}. \citeauthor{lingtea} argue that relying solely on English data leads to insufficient forgetting if the target knowledge has been acquired from multiple languages. \citeauthor{hwang2025uncovering} report the rise of language confusion from English-centric unlearning, while concurrent work by \citeauthor{lau} demonstrates the occurrence of cross-linguistic spread of sensitive information across languages. All three works suggested multilingual parallel unlearning as the solution. However, evaluations in these works are largely limited to a direct extension of English-centric protocols relying solely on per-language evaluation. It is questionable whether they are sufficient to fully capture the complex multilingual characteristics of MMU.

As illustrated in Figure~\ref{fig:teaser}, current evaluation protocols can be misleading: a model may appear to have unlearned information in the evaluated language while the same knowledge remains accessible in another. Consequently, language-wise evaluations cannot determine whether the underlying information has truly been removed, and may overstate unlearning effectiveness. Reliable evaluation therefore requires metrics that verify information inaccessibility consistently across all languages.

In this paper, we establish a comprehensive MMU scenario by 1) suggesting how knowledge should be defined in multilingual setting and 2) clarifying the two distinct mechanisms for its acquisition. Upon this scenario, 3) we finally design two suitable metrics for multilingual evaluation. We identify knowledge in MMU as an instance that has been obtained and expressed in multiple languages. This knowledge can be attained by either direct memorization or indirect cross-linguistic spread. To simulate both settings, we generated a multilingual parallel dataset spanning 10 languages, each containing 3,800 instances, where eight languages are used for memorization while the others are held out for evaluation. We assessed both scenarios using our metrics designed to capture the multilingual nature of knowledge: the Knowledge Separability Score (KSS), which evaluates the overall unlearning quality across all languages, and the Knowledge Persistence Score (KPS), which specifically quantifies consistent removal of information between language pairs. Through the extensive evaluations, we provide deeper insights into the unique phenomena of MMU, and present a new paradigm for evaluation.

To sum up, our contributions are as follows:
\begin{itemize}
    \item We conducted extensive analysis and experiments on various unlearning methods. To this end, we construct a large-scale multilingual parallel dataset (3,800 QA $\times$ 10 Languages).
    \item We proposed Knowledge Separability Score (KSS) and Knowledge Persistence Score (KPS) to evaluate the performance in MMU.
    \item Through extensive analysis using KSS and KPS, we demonstrate the usefulness of specialized metrics tailored for accurately measuring performance in MMU.
\end{itemize}

\section{Related Work}

\subsection{Machine Unlearning}

Machine Unlearning (MU) aims to selectively eliminate sensitive information from pre-trained LLMs while preserving the remaining knowledge. Existing approaches are typically grouped into optimization-based methods~\cite{ga, grad_diff, npo} and pruning-based methods~\cite{pochinkov2024dissecting}. However, existing studies on MU have been largely English-centric, which is misaligned with the multilingual nature of modern LLM deployment. Multilingual MU (MMU) studies~\cite{lingtea, hwang2025uncovering, lau} have emerged under such context, pointing out the insufficiency of English-only unlearning. They have analysed unique phenomena and developed unlearning methods, yet, effective evaluation of multilingual unlearning performance remains unexplored.

\subsection{Evaluation}
Up until now, MU evaluation protocols have largely been developed in English-centric settings. Existing metrics can be broadly categorized into two groups: 1) probability-based metrics and 2) generation-based metrics. Probability-based metrics assess how confidently a model knows the information. For example, TOFU~\cite{tofu} uses the probabilities assigned to the corresponding answer to quantify degrees of forgetting and retention. In contrast, generation-based metrics \hj{either} measure output-level agreement with a reference~\cite{lin-2004-rouge} or rely on LLM-as-a-judge style evaluations~\cite{liu2025protecting}.

These protocols are frequently applied to MMU without modification~\cite{lingtea, hwang2025uncovering}. However, 
MMU scenario is different from English-centric scenario in two aspects. First, knowledge is not confined to a single language but is distributed across multiple languages. Second, such multilingual information is acquired through both direct memorization~\cite{lingtea} and indirect cross-linguistic spread~\cite{lau}. On this viewpoint, we identified two limitations of applying English-centric evaluations directly to MMU: 1) evaluating each language in isolation is insufficient to verify whether specific information has been completely removed across the entire languages, and 2) existing researches typically address only one of the two knowledge acquirement mechanisms. To this end, we propose two metrics that can evaluate knowledge across multiple languages, and conduct experiments on both settings within a unified framework.
\section{Problem Formulation} \label{sec:problem_formulation}

\paragraph{Multilingual MU} \hj{In Machine Unlearning (MU), there are three states of a model. A \textit{pre-trained model}, $F_{\theta_0}$, refers to the model that has not yet been fine-tuned on specific dataset. After being fine-tuned to memorize specific information, the model becomes a \textit{memorized model} denoted as $F^M_\theta$. Finally, the \textit{unlearned model} that has been updated to forget some memorized knowledge is denoted as $F^U_{\theta}$.} \km{For MU tasks, three types of datasets are required: a fine-tuning set $\mathcal{D}$, a forget set $\mathcal{D}_f$, and a retain set $\mathcal{D}_r$. For MMU tasks, all three datasets $\mathcal{D}_f$, $\mathcal{D}_r$, and $\mathcal{D}$ \nj{consist} of multilingual parallel QA pairs, where each pair contains semantically equivalent content across different languages:}
\begin{equation}
\begin{aligned}
\mathcal{D}&=\{k_{i,l} \triangleq (q_{i,l}, a_{i,l})\:| \: i\in \mathcal{I}, \: l \in \mathbb{L}\},\\
\mathcal{D}_f&=\{k_{i,l} \triangleq(q_{i,l}, a_{i,l})\:| \: i\in \mathcal{I}_f, \: l \in \mathbb{L}\},\\
\mathcal{D}_r&=\{k_{i,l} \triangleq (q_{i,l}, a_{i,l})\:| \: i\in \mathcal{I}_r, \: l \in \mathbb{L}\},
\end{aligned}
\end{equation}
\km{where $\mathbb{L}$ denotes the set of languages \hj{and} $k_{i,l}$ indicates the $i$-th instance in language $l$. $\mathcal{I}$ is the union of two disjoint index sets $\mathcal{I}_f$ and $\mathcal{I}_r$, each enumerating the instances in the forget set and the retain set ($\mathcal{I} = \mathcal{I}_f \cup \mathcal{I}_r, \   \mathcal{I}_f \cap \mathcal{I}_r = \emptyset$). Similarly, $\mathcal{D}$ denotes the disjoint union of $\mathcal{D}_f$ and $\mathcal{D}_r$.} \hj{Dataset $\mathcal{D}$ can be viewed as a two dimensional $|\mathcal{I}|\times|\mathbb{L}|$ matrix with index-wise rows and language-wise columns.}

\km{Unlearning methods commonly employ the following loss function \nj{on top of $F_\theta^M$}:}
%
\begin{equation}
\mathcal{L}(\mathcal{D}_{f}, \mathcal{D}_r) = \mathcal{L}_f(\mathcal{D}_f) + \mathcal{L}_r(\mathcal{D}_r),
\end{equation}
%
\nj{where} \hj{$\mathcal{L}_f$ and $\mathcal{L}_r$ denotes the forget and retain loss.} 

\begin{figure}[t]
    \centering
    \includegraphics[width=1.0\linewidth]{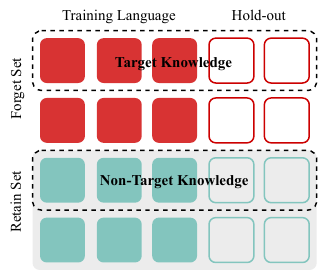}
    \caption{\hj{Overview illustration of our setting. A knowledge refers to an instance which may be expressed multilingually. Target Knowledge is the knowledge in the forget set, while Non-Target Knowledge is the one in the retain set.} In this setting, we propose metrics specifically designed for the evaluation of the knowledge.}
    \label{fig:setting}
\end{figure}

\paragraph{Our Setting} \km{In traditional English-centric MU, knowledge is expressed solely in English. However, unlike this English-centric approach, knowledge in Multilingual MU (MMU) can be expressed across multiple languages. Such knowledge can be acquired directly through multilingual training or derived from cross-lingual spread. For MMU, we categorized knowledge into \textit{Target Knowledge} (to be unlearned) and \textit{Non-target Knowledge} (to be retained). \nj{With abuse of Matlab matrix notation,} we formally define the $i$-th Target Knowledge ($k_i^\text{T}$) and $j$-th Non-target Knowledge ($k_j^\text{N}$) as } 
\begin{equation}
 k_i^\text{T} = k_{i,:}, \  i \in \mathcal{I}_f,  \quad k_j^\text{N} = k_{j,:}, \ j \in \mathcal{I}_r.
\end{equation}

\begin{figure*}[t]
    \centering
    \includegraphics[width=0.95\linewidth]{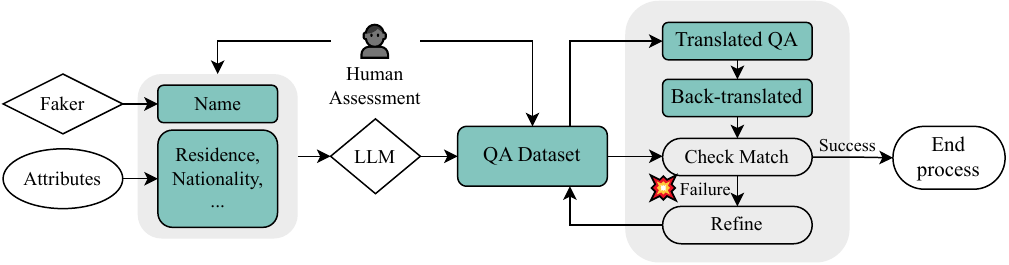}
    \caption{\km{Overview of }\hj{multilingual parallel QA dataset generation pipeline}}
    \label{fig:pipeline}
\end{figure*}

\km{\hj{Each $k_i^\text{T}$ and $k_j^\text{N}$} is composed of various languages but shares identical semantics. In this context, MMU must remove the target knowledge while retaining the non-target knowledge.}

\km{In Multilingual LLMs, knowledge acquired in one language spreads to other languages, a phenomenon denoted as cross-linguistic spread~\cite{lau}. To measure \nj{the unlearning performance} in this context, we conducted experiments using a setting that includes hold-out languages that were \nj{not} utilized in \nj{either} the memorization \nj{or} the unlearning phases. }\hj{To simulate this scenario, we employed $10$ languages. Five were chosen from high-resource languages:  \textsc{English}, \textsc{Chinese}, \textsc{German}, \textsc{Russian} and \textsc{Spanish}, while the others were chosen from low-resource languages: \textsc{Bengali}, \textsc{Hebrew}, \textsc{Tamil}, \textsc{Afrikaans} and \textsc{Albanian}.}

\hj{The selected languages are divided into \textit{Training} and \textit{Hold-out} languages for observation. The training languages are directly utilized for memorization and unlearning, while hold-out languages are only employed during evaluation.}
\vspace{-2mm}
\begin{itemize}
    \item \hj{Training}: \textsc{English, Chinese, German, Russian, Bengali, Hebrew, Tamil, Albanian}
    \item \hj{Hold-out}: \textsc{Afrikaans, Spanish}.
\end{itemize}

\hj{Here, we denote the set of training languages and hold-out languages as $\mathbb{L}_\text{Train}$ and $\mathbb{L}_\text{Hold}$, \km{respectively}. Figure~\ref{fig:setting} summarizes our overall setting.}

\section{Dataset Generation}

\subsection{Overview} \label{sec_dataset_generation}

\km{Knowledge within multilingual LLMs is often distributed across diverse languages instead of being confined to a single linguistic context. To simulate such setting, we introduced a multilingual parallel dataset. Inspired by TOFU~\cite{tofu}, we first generated 200 synthetic profiles \hj{to clearly isolate the effect of unlearning from the model's pre-trained knowledge.} From the profiles, an English question-answer (QA) dataset was constructed with 19 attribute-specific QA pairs. We subsequently translated the English QA dataset into \hj{9 other} languages (\hj{four} high-resource languages and five low-resource languages) to conduct MMU experiments. Figure~\ref{fig:pipeline} demonstrates the overview of data-generation pipeline: 1) generate English synthetic profiles, 2) prompt an LLM to produce English QA pairs for each profile, 3) translate the QA pairs into multiple languages to form a parallel multilingual MMU dataset, and 4) verify the translations via back-translation \hj{to English}.}

\subsection{Synthetic Profile Generation}

\km{We \hj{assigned} 20 attributes \hj{to} each synthetic \hj{profile}, including \textsc{Name}, \textsc{Year of Birth} and etc. Before constructing the attributes, we generated 200 unique fictitious names in English using the Faker~\cite{faker} library. We then pre-specified values for every attribute. The value pools for each attribute are listed in Appendix~\ref{sec:att_pool}. To improve the quality of the synthetic profiles, human annotators reviewed the generated profiles and removed cases that were inconsistent with common sense (e.g., a barista working fully remote). Examples of the resulting profiles appear in Appendix~\ref{sec:profile_ex}.}

\subsection{QA Dataset Generation}

\km{We employed the Qwen3-225B-A22B-Thinking-2507~\cite{qwen3} model to generate 19 distinct QA datasets from the 200 synthetic profiles introduced in the previous section. To make each question focus on a single attribute, we provide the LLM with only the subject’s name and one attribute, then have it generate the corresponding QA pair. The full prompt is provided in Figure~\ref{fig:pool_prompt_qa}. For quality control, human annotators manually corrected each QA pair whose content \hj{is not aligned with the} corresponding profile. Representative QA examples appear in Figure~\ref{fig:syn_ex}.}

\subsection{Translate QA Dataset}

\begin{figure*}[t]
    \centering
    \includegraphics[width=1.0\linewidth]{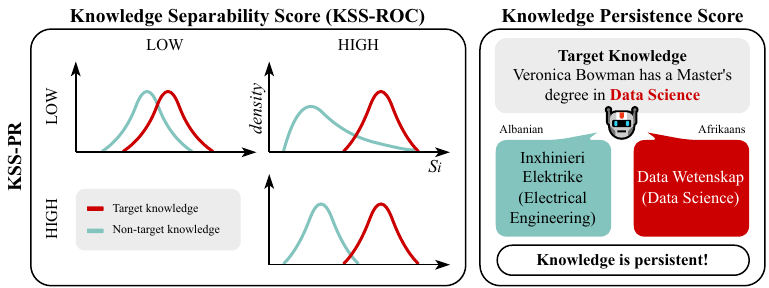}
    \caption{\km{Overview of the Knowledge Separability Score (KSS) and Knowledge Persistence Score (KPS). KSS-ROC measures the overall separability between the target and non-target knowledge, while KSS-PR evaluates how consistently the model assigns higher $S_i$ to the target knowledge compared to the non-target knowledge. KPS quantifies the extent to which knowledge inaccessible in one language but persists in another.}}
    \label{fig:metric_fig}
\end{figure*}

\km{We translated the 3,800 English QA pairs (19$\times$200), derived from synthetic profiles, into 9 languages using the Google Translation API~\cite{google-translate}. Because the profiles are synthetic and thus unfamiliar to the models, maintaining identity consistency across languages is crucial. Accordingly, following prior multilingual benchmarks~\cite{wikiann,wikimatrix}, we leave personal names untranslated.} \km{To ensure translation quality, we adopt a back-translation-based verification-and-refinement pipeline inspired by \citet{joshi2025cultureguard}. Specifically, we employed Google Translation API to translate each English source sentence into the target language, and back into English. Then we assess semantic equivalence using Qwen3-225B-A22B-Thinking-2507 (see Figure~\ref{fig:back_translated_sentence_verify_prompt} for the verification prompt). If equivalence check fails, we revise it using ChatGPT~\cite{gpt} and repeat the above process. We iterate this \hj{verify-and-refine} cycle until a human annotator confirms that the back-translation is semantically equivalent to the source English dataset. We apply this procedure to all target languages for semantically consistent translations. Examples from the resulting multilingual parallel QA dataset are shown in Figure~\ref{fig:qa_ex_multi}.}

\section{Knowledge Evaluation in MMU} \label{sec:metric}

\km{Current MMU evaluation methods are typically direct extensions of English-centric approaches conducted in a language-wise manner. However, this approach fails to wholly assess knowledge distributed across diverse languages. To address such limitation, we proposed new metrics based on the following principles: 1) \textit{Holistic Evaluation of Unlearning Quality}: Measuring MMU performance requires a unified metric capable of assessing knowledge across multiple languages. 2) \textit{Cross-lingual Consistent Forgetting}: Metrics should specifically quantify the consistent removal of sensitive information between language pairs. To this end, we first review existing metrics used in language-wise evaluation and then propose novel metrics specifically tailored for MMU. Figure~\ref{fig:metric_fig} provides an overview of the aspects measured by our proposed metrics.}

\subsection{Language-wise Evaluation} \km{Prior MMU studies directly extend the English-centric protocols, conducting performance evaluations separately \nj{for each} language. Commonly used evaluation metrics are broadly categorized into two types: probability-based and generation-based.}
    
\paragraph{Probability} \hj{is} \km{measured \hj{as} the conditional probability
$\mathcal{P}(a \mid q)^{1/|a|_{\text{tok}}}$, where $q$ denotes the question sentence and
$a$ denotes the corresponding answer. $|a|_{\text{tok}}$ is the token length of $a$.}

\paragraph{Semantic Equivalence} \hj{LLMs can} assess whether the model outputs are semantically identical to the ground truth. To mitigate potential ambiguity arising from evaluating in low-resource languages, we translated the generated outputs and the ground truths into English using NLLB-200-3.3B, a multilingual model specialized in translation~\cite{nllb}. \hj{We define Semantic Equivalence (SE) as follows:}

\begin{equation}
\text{SE}(q, a) = \mathbb{I} \left( \text{LLM}(\mathcal{T}(F_\theta(q)), \mathcal{T}(a)) 
\right), 
\end{equation}

\noindent where $\mathcal{T}$ denotes the \nj{translation into English}, and $\mathbb{I}(\text{LLM}(\cdot, \cdot))$ outputs 1 if the LLM determines that the two inputs have the same meaning, and 0 if they do not. We employed GPT-4o-mini~\cite{achiam2023gpt} with greedy decoding for semantic equivalence judgement. The prompt used for evaluation is provided in Figure~\ref{fig:prompt_for_ser}. Previous MMU researches utilized such SE score to evaluate the knowledge of LLM in a language-wise manner.

\subsection{Knowledge \nj{(Instance)}-wise Evaluation}
Existing MU metrics cannot adequately assess unlearning performance in MMU: \emph{these metrics fail to capture properties that arise uniquely in multilingual scenarios}. To address this limitation, we propose two knowledge-wise metrics: (1) the Knowledge Separability Score (KSS), which summarizes the unlearning performance for both target and non-target knowledge and (2) the Knowledge Persistence Score (KPS), which quantifies the extent to which a target knowledge that is inaccessible in one language remains retrievable in another.

\paragraph{Knowledge Separability Score} \km{We proposed the Knowledge Separability Score (KSS) as a comprehensive AUC-based measure for MMU performance. KSS is computed in two steps: 1) we derive a knowledge-wise forgetting score $S_i$ that quantifies the degree of forgetting for the $i$-th knowledge, and 2) we use these scores to compute the AUC over the forget and retain sets.}

\km{\hj{We calculate the knowledge-wise forgetting score $S_i$ for the $i$-th QA pair across $\mathbb{L}$ languages, $\{(q_{i,l}, a_{i,l}) \ | \ l\in \mathbb{L}\}$, in two ways.} First, the generation-based score $S^{gen}_i$ aggregates the Semantic Equivalence (SE) in a knowledge-wise manner and quantifies the inequivalence by subtracting it from $1$. Second, the probability-based score $S^{prob}_i$ utilizes the length-normalized probability assigned to the ground truth sequence, $\mathcal{P}(a_{i, l} | q_{i,l})^{1/|a_{i,l}|_{\text{tok}}}$. We subtract it from $1$ so that a lower $\mathcal{P}$ corresponds to a higher $S_i$. The scores are formally defined as:}
\begin{equation}
\begin{aligned}
    S^{gen}_i &= 1-\frac{1}{|\mathbb{L}|}\sum_{l \in \mathbb{L}}\text{SE}(q_{i, l}, a_{i, l}), \\
    S^{prob}_i &= 1 - \frac{1}{|\mathbb{L}|} \sum_{l \in \mathbb{L}} \mathcal{P}(a_{i, l} | q_{i, l})^{1/|a_{i, l}|_{tok}}.
\end{aligned}
\end{equation}

\km{We computed $S_i$ for both target and non-target knowledge and plotted the probability density functions, as shown in Figure~\ref{fig:metric_fig}.} \km{Using these functions, we measure KSS using two complementary metrics: \nj{Area Under the Receiver Operating Characteristic Curve (KSS-ROC) and that of the Precision-Recall Curve (KSS-PR)\footnote{\nj{Note that ROC is drawn based on TPR (true positive ratio = TP/(TP+FN); y-axis) vs. FPR (false positive ratio = FP/(FP+TN); x-axis), while PR is drawn from Precision (TP/(TP+FP); y-axis) and Recall (TP/(TP+FN); x-axis).}}}. We computed KSS-ROC and KSS-PR by varying the threshold for $S_i$.}

\km{While KSS-ROC provides a general measure of separability \nj{between forget (target) and retain (non-target) sets}, KSS-PR further addresses the severe forget-retain dataset imbalance, i.e., the forget set is typically much smaller than the retain set. Specifically, a high KSS-ROC signifies that the $S_i$ distributions between the forget and retain sets are effectively distinguishable, whereas a high KSS-PR suggests that the model yields consistently elevated $S_i$ scores for the forget dataset.}

\km{Both KSS-ROC and KSS-PR are indispensable metrics for the precise evaluation. As demonstrated in Figure~\ref{fig:metric_fig}, a high KSS-ROC score alone does not guarantee that non-target knowledge is free from erroneously assigned high forgetting scores ($S_i$). Conversely, a low KSS-PR score does not necessarily imply a lack of global separability. Therefore, these two metrics are mutually complementary. We provide the detailed explanation in Appendix~\ref{sec:det_metric}.}

\paragraph{Knowledge Persistence Score} \km{To quantify the degree of persistence of the target knowledge, we proposed the Knowledge Persistence Score (KPS). For a base language $l_1$ and a comparison language $l_2$, we define the pairwise persistence score as the fraction of samples that are judged as forgotten in $l_1$ ($\text{SE}(q_{i, l_1}, a_{i, l_1})=0$) but still retained in $l_2$ ($\text{SE}(q_{i, l_2}, a_{i, l_2})=1$):}

\begin{equation}
\begin{aligned}
ps(l_1, l_2)&=\frac{1}{|\mathcal{I}(l_1)|}\sum_{i\in \mathcal{I}(l_1)} \text{SE}(q_{i,l_2}, a_{i,l_2}),\\
\mathcal{I}(l_1)&\triangleq\{i \in \mathcal{I}_f\ | \ \text{SE}(q_{i, l_1}, a_{i, l_1})=0\}.
\end{aligned}
\label{eq:ps_small}
\end{equation}
\km{$ps(l_1, l_2)$ is the retention \hj{of the target knowledge} in $l_2$ conditioned on forgetting in $l_1$. Specifically, it serves to measure how consistently the forgetting occurs between the languages.}

\km{Given a set of comparison languages $\mathbb{L}_2 \: \text{s.t.} \: l_1 \notin \mathbb{L}_2$, we aggregate pairwise persistence scores by averaging over $l_2 \in \mathbb{L}_2$:}

\begin{equation}
\text{KPS}(l_1, \mathbb{L}_2)=\frac{1}{|\mathbb{L}_2|}\sum_{l_2 \in \mathbb{L}_2}ps(l_1, l_2).
\end{equation}

\km{KPS provides a quantitative measure of how easily the target knowledge, once unlearned in $l_1$, can be recovered by querying the model in $\mathbb{L}_2$.}
\nj{A small value of KPS represents better unlearning performance in MMU. }

\subsection{Experimental Setting}

\begin{table*}[t]
    \centering
    \scriptsize 
    \renewcommand{\arraystretch}{1.25}
    \setlength{\tabcolsep}{3pt}
    
    \newcommand{\res}[2]{\textbf{#1}$_{\scriptscriptstyle +#2}$}
    \newcommand{\base}[1]{#1}

    \resizebox{\textwidth}{!}{%
    \begin{tabular}{ll | cc cc cc || cc cc cc}
        \toprule
        \multirow{3}{*}{\textbf{Method}} & \multirow{3}{*}{\textbf{Type}} & \multicolumn{6}{c||}{\textbf{KSS-ROC} ($\uparrow$)} & \multicolumn{6}{c}{\textbf{KSS-PR} ($\uparrow$)} \\
        \cmidrule(lr){3-8} \cmidrule(lr){9-14}
        
         & & \multicolumn{2}{c}{\textbf{p1}} & \multicolumn{2}{c}{\textbf{p3}} & \multicolumn{2}{c||}{\textbf{p5}} & \multicolumn{2}{c}{\textbf{p1}} & \multicolumn{2}{c}{\textbf{p3}} & \multicolumn{2}{c}{\textbf{p5}} \\
         & & \textbf{Case 1} & \textbf{Case 2} & \textbf{Case 1} & \textbf{Case 2} & \textbf{Case 1} & \textbf{Case 2} & \textbf{Case 1} & \textbf{Case 2} & \textbf{Case 1} & \textbf{Case 2} & \textbf{Case 1} & \textbf{Case 2} \\
        \midrule

        \multirow{2}{*}{\textbf{MEM}} 
          & Prob & \base{0.52} & \base{0.45} & \base{0.51} & \base{0.49} & \base{0.51} & \base{0.49} & \base{0.01} & \base{0.01} & \base{0.03} & \base{0.03} & \base{0.05} & \base{0.05} \\
          & Gen  & \base{0.51} & \base{0.50} & \base{0.51} & \base{0.48} & \base{0.51} & \base{0.49} & \base{0.01} & \base{0.03} & \base{0.03} & \base{0.03} & \base{0.05} & \base{0.05} \\
        \cmidrule{1-14}

        \multirow{2}{*}{\textbf{GA}} 
          & Prob 
          & \res{0.57}{10} & \res{0.89}{98} & \res{0.53}{4} & \res{0.81}{65} & \res{0.52}{2} & \res{0.66}{35} 
          & \res{0.01}{0} & \res{0.39}{3800} & \res{0.04}{33} & \res{0.24}{700} & \res{0.05}{0} & \res{0.12}{140} \\
          & Gen  
          & \res{0.57}{12} & \res{0.70}{40} & \res{0.53}{4} & \res{0.65}{35} & \res{0.53}{4} & \res{0.55}{12} 
          & \res{0.01}{0} & \res{0.15}{400} & \res{0.03}{0} & \res{0.10}{233} & \res{0.05}{0} & \res{0.07}{40} \\
        \cmidrule{1-14}

        \multirow{2}{*}{\textbf{GAGDR}} 
          & Prob 
          & \res{0.61}{17} & \res{0.91}{102} & \res{0.54}{6} & \res{0.78}{59} & \res{0.54}{6} & \res{0.72}{47}
          & \res{0.02}{100} & \res{0.46}{4500} & \res{0.03}{0} & \res{0.14}{367} & \res{0.06}{20} & \res{0.15}{200} \\
          & Gen  
          & \res{0.57}{12} & \res{0.77}{54} & \res{0.52}{2} & \res{0.65}{35} & \res{0.52}{2} & \res{0.62}{27}
          & \res{0.01}{0} & \res{0.18}{500} & \res{0.03}{0} & \res{0.05}{67} & \res{0.05}{0} & \res{0.08}{60} \\
        \cmidrule{1-14}

        \multirow{2}{*}{\textbf{GAKLR}} 
          & Prob 
          & \res{0.66}{27} & \res{0.96}{113} & \res{0.57}{12} & \res{0.83}{69} & \res{0.55}{8} & \res{0.71}{45}
          & \res{0.02}{100} & \res{0.64}{6300} & \res{0.04}{33} & \res{0.20}{567} & \res{0.07}{40} & \res{0.13}{160} \\
          & Gen  
          & \res{0.67}{31} & \res{0.85}{70} & \res{0.56}{10} & \res{0.69}{44} & \res{0.55}{8} & \res{0.62}{27}
          & \res{0.02}{100} & \res{0.47}{1467} & \res{0.03}{0} & \res{0.10}{233} & \res{0.06}{20} & \res{0.10}{100} \\
        \cmidrule{1-14}

        \multirow{2}{*}{\textbf{NPO}} 
          & Prob 
          & \res{0.70}{35} & \res{0.99}{120} & \res{0.59}{16} & \res{0.89}{82} & \res{0.51}{0} & \res{0.65}{33}
          & \res{0.03}{200} & \res{0.88}{8700} & \res{0.06}{100} & \res{0.53}{1667} & \res{0.05}{0} & \res{0.17}{240} \\
          & Gen  
          & \res{0.66}{29} & \res{0.91}{82} & \res{0.60}{18} & \res{0.74}{54} & \res{0.56}{10} & \res{0.59}{20}
          & \res{0.02}{100} & \res{0.48}{1500} & \res{0.04}{33} & \res{0.19}{533} & \res{0.06}{20} & \res{0.08}{60} \\
        \cmidrule{1-14}

        \multirow{2}{*}{\textbf{PRUNE}} 
          & Prob 
          & \res{0.76}{46} & \res{0.91}{102} & \res{0.66}{29} & \res{0.85}{73} & \res{0.63}{24} & \res{0.82}{67}
          & \res{0.07}{600} & \res{0.12}{1100} & \res{0.06}{100} & \res{0.16}{433} & \res{0.08}{60} & \res{0.18}{260} \\
          & Gen  
          & \res{0.68}{33} & \res{0.90}{80} & \res{0.68}{33} & \res{0.83}{73} & \res{0.62}{22} & \res{0.79}{61}
          & \res{0.02}{100} & \res{0.08}{167} & \res{0.05}{67} & \res{0.15}{400} & \res{0.07}{40} & \res{0.15}{200} \\
        \bottomrule
    \end{tabular}%
    }
    \caption{Performance of KSS-ROC and KSS-PR scores of various unlearning methods. Subscripts denote the percentage increase relative to MEM (e.g., $0.57_{+10}$ means 10\% increase). MEM denotes the memorized model ($F_\theta^M$), Prob denotes the probability-based scores and Gen denotes the generation-based scores.}
    \label{tab:combined_clean}
\end{table*}

\paragraph{Unlearning Configuration} \km{We employed a set of widely used optimization-based unlearning algorithms--Gradient Ascent (GA)~\cite{ga}, Gradient Ascent with Gradient Descent term (GAGDR)~\cite{grad_diff}, Gradient Ascent with KL minimization (GAKLR)~\cite{tofu} and Negative Preference Optimization (NPO)~\cite{npo}. Additionally, we conducted experiments using a pruning-based unlearning method~\cite{pochinkov2024dissecting}. Detailed descriptions are provided in Appendix~\ref{unlearn_method_detail}.}

\km{We conducted experiments using Llama3.1-8B-Instruct (Llama3.1), a multilingual LLM, as the base model~\cite{llama3}. We used the multilingual parallel QA dataset described in Section~\ref{sec_dataset_generation} for both fine-tuning (memorization) and unlearning across all methods. We considered the $p1$, $p3$, and $p5$ settings according to the ratio of the forget set (1\%, 3\% and 5\% respectively). Detailed hyperparameters are provided in Appendix~\ref{hyperparam_setting}.} 

\paragraph{Evaluation Configuration} In the multilingual unlearning scenario, knowledge is acquired either through direct training ($\mathbb{L}_\text{train}$) or cross-linguistic spread ($\mathbb{L}_\text{Hold}$). Since the two subsets have acquired knowledge in different ways, it is more adequate to analyse both KSS and KPS on $\mathbb{L}_\text{train}$ and $\mathbb{L}_\text{Hold}$ each instead of aggregating the two.
\section{Analysis}

\subsection{Knowledge Separability Score} \label{sec:kss_detail}

\hj{We reported KSS of two cases: }
\begin{itemize}
    \item \textbf{Case 1}: The separability between target and non-target knowledge within $\mathbb{L}_\text{Hold}$,
    \item \textbf{Case 2}: The separability between target and non-target knowledge within $\mathbb{L}_\text{Train}$.
\end{itemize}

\paragraph{Unlearning is Difficult in Hold-out Languages (Case 1)} \km{We observed a distinct performance disparity between Case 1 and Case 2. Regarding KSS-ROC in Table~\ref{tab:combined_clean}, scores are consistently lower for Case 1 (measured within $\mathbb{L}_\text{Hold}$) compared to Case 2 (measured within $\mathbb{L}_\text{Train}$) for both probability- and generation-based metrics. For example, for $p1$, the maximum score is 0.99 in Case 2, whereas it reaches only 0.76 in Case 1. This indicates that distinguishing between target and non-target knowledge is more challenging in Case 1.}

\paragraph{Unlearning Performance Degrades as Forget Ratio Increases (Case 2)} Table~\ref{tab:combined_clean} presents the performance of probability- and generation-based KSS scores measured by ROC-AUC (KSS-ROC) and PR-AUC (KSS-PR). To ensure a fair comparison given the performance variability of the Memorized model (MEM) across different forget dataset ratios, we report the absolute scores alongside the percentage increase relative to MEM. The relative increase is calculated as $\frac{\text{Method} - \text{MEM}}{\text{MEM}} \times 100$ and is denoted by a subscript (e.g., $_{+10}$). As shown in the table, the performance of both metrics degrades as the forget dataset ratio increases from $p1$ to $p5$, except for KSS-PR in PRUNE. This suggests that as the forget ratio increases, the boundary between target and non-target knowledge becomes increasingly obscure, making it difficult for the model to distinguish between the two.

\paragraph{Analysis on Prune-based Method (Case 2)} The prune-based method demonstrates high KSS-ROC score in the $p1$ setting within $\mathbb{L}_\text{Train}$ (Case 2), where unlearning is generally effective across all methods. This indicates that, like optimization-based methods, pruning can successfully achieve strong global separability between target and non-target knowledge. However, we also observe that its KSS-PR score is disproportionately poor, remaining significantly lower than that of other methods with comparable KSS-ROC scores.

To investigate the cause of this discrepancy, we visualized the distributions of knowledge-wise forgetting scores ($S_i$) for both the optimization-based methods and the prune-based method under the $p1$ setting (Figure~\ref{fig:clfs_analysis_main}). The distributions for all forget ratios and methods are provided in Appendix~\ref{sec:kps_detail}. From the visualization, we found that the pruned model has assigned high $S_i$ to not only the target knowledge, but also to non-negligible amount of the non-target knowledge. In other words, pruning has failed to assign sufficiently distinct, high knowledge-wise forgetting scores exclusively to the target knowledge, relatively to optimization-based methods. This results in a significant overlap of target knowledge with the tail of the non-target knowledge distribution (highlighted by the red box). Consequently, this leads to a degradation in KSS-PR, indicating that target knowledge does not exclusively reside in the high-score region.

\begin{figure}[t]
    \centering
    \includegraphics[width=0.95\linewidth]{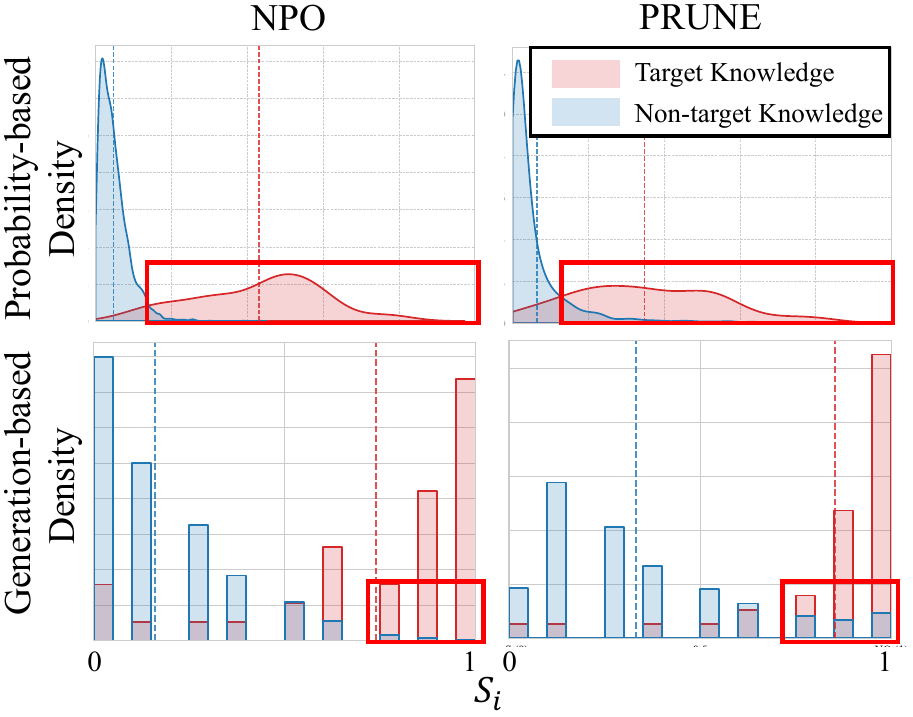}
    \caption{\km{Distributions of $S_i$ for both the target and non-target knowledge after NPO and pruning in Case 2. The first row represents the probability-based $S_i$, while the second row displays the generation-based $S_i$.}}
    \label{fig:clfs_analysis_main}
\end{figure}

\subsection{Knowledge Persistence Score} \label{sec:kps_detail}

\hj{We now report KPS of two cases: }
\begin{itemize}
    \item \textbf{Case 1}: Target knowledge inaccessible in the base language $l_1\in \mathbb{L}_\text{Train}$, but still persists within $\mathbb{L}_\text{Hold}$,
    \item \textbf{Case 2}: Target knowledge inaccessible in the base language $l_1\in \mathbb{L}_\text{Train}$, but still persists within $\mathbb{L}_\text{Train}\setminus \{l_1\}$.
\end{itemize}

\paragraph{Knowledge Can Persist in Hold-out Languages (Case 1)}
\hj{While it is straightforward that more knowledge persists in Case 2, the results of Case 1 show that cross-linguistic spread of knowledge persists in hold-out languages even after unlearning. For every base language $l_1$ utilized for the measurement, there exists unremoved target knowledge to the hold-out languages ($\text{KPS}>0$). This phenomenon again raise the potential risk of \km{cross-lingual persistence} in MMU.} 

\paragraph{Persistence Tendency in Forget Set (Case 1 \& Case 2)}
\hj{Table~\ref{tab:kps_npo_performance} displays the unlearning performance of NPO measured with the Knowledge Persistence Score \km{(KPS)}. As in KSS, KPS also depicts more severe knowledge persistence as the forget ratio rises. Across every base language $l_1$, $p1$ setting shows the lowest KPS score that ranges from $\text{KPS}=0.05$ at the lowest and $\text{KPS}=0.18$ at the highest. On the other hand, $p3$ and $p5$ displays severe persistence with up to $\text{KPS}=0.44$. This implies that, as the proportion of forget set increases, unlearning becomes more difficult in the perspective of consistent unlearning between languages.}

\begin{table}[t]
\centering
\resizebox{\columnwidth}{!}{%
    \begin{tabular}{lcccccc}
    \toprule
    \multicolumn{7}{c}{\textbf{KPS ($\downarrow$)}} \\
    \midrule
    \multirow{2}{*}{$l_1$} & \multicolumn{2}{c}{\textbf{$p1$}} & \multicolumn{2}{c}{\textbf{$p3$}} & \multicolumn{2}{c}{\textbf{$p5$}} \\
    \cmidrule(lr){2-3} \cmidrule(lr){4-5} \cmidrule(lr){6-7}
     & \textbf{Case 1} & \textbf{Case 2} & \textbf{Case 1} & \textbf{Case 2} & \textbf{Case 1} & \textbf{Case 2} \\
    \midrule
    bn             & 0.08 & 0.17 & 0.20 & 0.41 & 0.19 & 0.36 \\
    de             & 0.08 & 0.15 & 0.10 & 0.37 & 0.14 & 0.30 \\
    en             & 0.05 & 0.07 & 0.05 & 0.17 & 0.05 & 0.17 \\
    he             & 0.09 & 0.14 & 0.11 & 0.44 & 0.17 & 0.36 \\
    ru             & 0.11 & 0.18 & 0.06 & 0.34 & 0.13 & 0.29 \\
    sq             & 0.11 & 0.13 & 0.09 & 0.42 & 0.17 & 0.26 \\
    ta             & 0.13 & 0.17 & 0.19 & 0.44 & 0.19 & 0.40 \\
    zh             & 0.13 & 0.15 & 0.16 & 0.40 & 0.16 & 0.32 \\
    \midrule
    \textbf{avg}   & \textbf{0.10} & \textbf{0.15} & \textbf{0.12} & \textbf{0.37} & \textbf{0.15} & \textbf{0.31} \\
    \bottomrule
    \end{tabular}%
}
\caption{Knowledge Persistence Score (KPS) on NPO across different forget ratios ($p1$, $p3$, $p5$).}
\label{tab:kps_npo_performance}
\end{table}

\section{Conclusion}

In this paper, we identified the operational unit of unlearning within Multilingual Machine Unlearning (MMU) and established a comprehensive evaluation protocols based on this new perspective. Leveraging a large-scale multilingual synthetic dataset constructed for this study, we conducted extensive experiments across various unlearning methods. To measure their performance regarding the multilingual characteristics, we introduced two metrics: the Knowledge Separability Score (KSS) and the Knowledge Persistence Score (KPS). These metrics enabled us to uncover and analyse unlearning dynamics unique to multilingual scenarios, providing deeper insights into the behavior of MMU. \hj{We conclude by suggesting that future research on MMU should \nj{consider multilingual characteristics} and aim to unlearn the knowledge across languages.}
\section{Limitations and Future Works}

\km{In this paper, we investigated unlearning performance evaluation within Multilingual Machine Unlearning (MMU) scenarios, where knowledge is distributed across diverse languages. To this end, we proposed the Knowledge Separability Score (KSS) and the Knowledge Persistence Score (KPS). Despite the contributions, our study has several limitations that suggest directions for future research.}

\km{First, there is a limitation regarding the diversity of training and hold-out languages. Although we selected a broad range of high- and low-resource languages across various language families to observe performance disparities, our scope for hold-out languages was restricted. Specifically, our experiments utilized only languages from the Indo-European family (i.e., Afrikaans and Spanish) as hold-out languages. Future research should incorporate a wider array of language families for the hold-out set to ensure a more comprehensive performance analysis across different linguistic structures.}

\km{Second, our experiments were limited by model scale. Due to computational constraints, we focused on an 8B-parameter model. We observed that effective unlearning was primarily achievable when the forget ratio was low; however, unlearning performance degraded significantly as the forget ratio increased. Since larger models may exhibit different behaviors regarding capacity and forgetting dynamics, it is crucial to validate these findings across a broader spectrum of model sizes.}

\km{Finally, while we proposed KSS and KPS with the consideration of knowledge-wise measurement in MMU contexts, there remains potential for alternative metrics. Future work should explore more diverse evaluation methodologies to verify the removal of knowledge more accurately and robustly.}

\section*{Acknowledgments}
This work was supported by the Korean Government through the grants from IITP (RS-2021-II211343, RS-2022-II220953, RS-2025-25442338).
%

\bibliography{custom}

@article{mu-survey,
  title={Machine unlearning: A comprehensive survey},
  author={Wang, Weiqi and Tian, Zhiyi and Zhang, Chenhan and Yu, Shui},
  journal={arXiv preprint arXiv:2405.07406},
  year={2024}
}

@article{ga,
  title={Knowledge unlearning for mitigating privacy risks in language models},
  author={Jang, Joel and Yoon, Dongkeun and Yang, Sohee and Cha, Sungmin and Lee, Moontae and Logeswaran, Lajanugen and Seo, Minjoon},
  journal={arXiv preprint arXiv:2210.01504},
  year={2022}
}

@article{npo,
  title={Negative preference optimization: From catastrophic collapse to effective unlearning},
  author={Zhang, Ruiqi and Lin, Licong and Bai, Yu and Mei, Song},
  journal={arXiv preprint arXiv:2404.05868},
  year={2024}
}

@article{tofu,
  title={Tofu: A task of fictitious unlearning for llms},
  author={Maini, Pratyush and Feng, Zhili and Schwarzschild, Avi and Lipton, Zachary C and Kolter, J Zico},
  journal={arXiv preprint arXiv:2401.06121},
  year={2024}
}

@article{muse,
  title={Muse: Machine unlearning six-way evaluation for language models},
  author={Shi, Weijia and Lee, Jaechan and Huang, Yangsibo and Malladi, Sadhika and Zhao, Jieyu and Holtzman, Ari and Liu, Daogao and Zettlemoyer, Luke and Smith, Noah A and Zhang, Chiyuan},
  journal={arXiv preprint arXiv:2407.06460},
  year={2024}
}

@article{gpt,
  title={Gpt-4 technical report},
  author={Achiam, Josh and Adler, Steven and Agarwal, Sandhini and Ahmad, Lama and Akkaya, Ilge and Aleman, Florencia Leoni and Almeida, Diogo and Altenschmidt, Janko and Altman, Sam and Anadkat, Shyamal and others},
  journal={arXiv preprint arXiv:2303.08774},
  year={2023}
}

@article{lingtea,
  title={Cross-lingual unlearning of selective knowledge in multilingual language models},
  author={Choi, Minseok and Min, Kyunghyun and Choo, Jaegul},
  journal={arXiv preprint arXiv:2406.12354},
  year={2024}
}

@inproceedings{grad_diff,
  title={Continual learning and private unlearning},
  author={Liu, Bo and Liu, Qiang and Stone, Peter},
  booktitle={Conference on Lifelong Learning Agents},
  pages={243--254},
  year={2022},
  organization={PMLR}
}

@article{qwen3,
  title={Qwen3 technical report},
  author={Yang, An and Li, Anfeng and Yang, Baosong and Zhang, Beichen and Hui, Binyuan and Zheng, Bo and Yu, Bowen and Gao, Chang and Huang, Chengen and Lv, Chenxu and others},
  journal={arXiv preprint arXiv:2505.09388},
  year={2025}
}

@misc{faker,
  author       = {Daniele Faraglia},
  title        = {Faker: Python package that generates fake data for you},
  year         = {2025},
  howpublished = {\url{https://github.com/joke2k/faker}}
}

@article{lau,
  title={Learn and unlearn in multilingual llms},
  author={Lu, Taiming and Koehn, Philipp},
  journal={arXiv preprint arXiv:2406.13748},
  year={2024}
}

@online{google-translate,
  author  = {Google Cloud},
  title   = {Cloud Translation documentation},
  year    = {2025},
  month   = {11},
  day     = {6},
  url     = {https://cloud.google.com/translate/docs},
  note    = {Accessed: 2025-11-11}
}

@article{joshi2025cultureguard,
  title={CultureGuard: Towards Culturally-Aware Dataset and Guard Model for Multilingual Safety Applications},
  author={Joshi, Raviraj and Paul, Rakesh and Singla, Kanishk and Kamath, Anusha and Evans, Michael and Luna, Katherine and Ghosh, Shaona and Vaidya, Utkarsh and Long, Eileen and Chauhan, Sanjay Singh and others},
  journal={arXiv preprint arXiv:2508.01710},
  year={2025}
}

@article{llama3,
  title={The llama 3 herd of models},
  author={Grattafiori, Aaron and Dubey, Abhimanyu and Jauhri, Abhinav and Pandey, Abhinav and Kadian, Abhishek and Al-Dahle, Ahmad and Letman, Aiesha and Mathur, Akhil and Schelten, Alan and Vaughan, Alex and others},
  journal={arXiv preprint arXiv:2407.21783},
  year={2024}
}

@inproceedings{wikiann,
    title = "Cross-lingual Name Tagging and Linking for 282 Languages",
    author = "Pan, Xiaoman  and
      Zhang, Boliang  and
      May, Jonathan  and
      Nothman, Joel  and
      Knight, Kevin  and
      Ji, Heng",
    editor = "Barzilay, Regina  and
      Kan, Min-Yen",
    booktitle = "Proceedings of the 55th Annual Meeting of the Association for Computational Linguistics (Volume 1: Long Papers)",
    month = jul,
    year = "2017",
    address = "Vancouver, Canada",
    publisher = "Association for Computational Linguistics",
    url = "https://aclanthology.org/P17-1178/",
    doi = "10.18653/v1/P17-1178",
    pages = "1946--1958"
}

@inproceedings{wikimatrix,
    title = "{W}iki{M}atrix: Mining 135{M} Parallel Sentences in 1620 Language Pairs from {W}ikipedia",
    author = "Schwenk, Holger  and
      Chaudhary, Vishrav  and
      Sun, Shuo  and
      Gong, Hongyu  and
      Guzm{\'a}n, Francisco",
    editor = "Merlo, Paola  and
      Tiedemann, Jorg  and
      Tsarfaty, Reut",
    booktitle = "Proceedings of the 16th Conference of the European Chapter of the Association for Computational Linguistics: Main Volume",
    month = apr,
    year = "2021",
    address = "Online",
    publisher = "Association for Computational Linguistics",
    url = "https://aclanthology.org/2021.eacl-main.115/",
    doi = "10.18653/v1/2021.eacl-main.115",
    pages = "1351--1361"
}

@inproceedings{lin-2004-rouge,
    title = "{ROUGE}: A Package for Automatic Evaluation of Summaries",
    author = "Lin, Chin-Yew",
    booktitle = "Text Summarization Branches Out",
    month = jul,
    year = "2004",
    address = "Barcelona, Spain",
    publisher = "Association for Computational Linguistics",
    url = "https://aclanthology.org/W04-1013/",
    pages = "74--81"
}

@article{hwang2025uncovering,
  title={Uncovering the Potential Risks in Unlearning: Danger of English-only Unlearning in Multilingual LLMs},
  author={Hwang, Kyomin and Kim, Hyeonjin and Kim, Seungyeon and Wee, Sunghyun and Kwak, Nojun},
  journal={arXiv preprint arXiv:2510.23949},
  year={2025}
}

@article{pochinkov2024dissecting,
  title={Dissecting language models: Machine unlearning via selective pruning},
  author={Pochinkov, Nicholas and Schoots, Nandi},
  journal={arXiv preprint arXiv:2403.01267},
  year={2024}
}

@article{vaswani2017attention,
  title={Attention is all you need},
  author={Vaswani, Ashish and Shazeer, Noam and Parmar, Niki and Uszkoreit, Jakob and Jones, Llion and Gomez, Aidan N and Kaiser, {\L}ukasz and Polosukhin, Illia},
  journal={Advances in neural information processing systems},
  volume={30},
  year={2017}
}

@article{achiam2023gpt,
  title={Gpt-4 technical report},
  author={Achiam, Josh and Adler, Steven and Agarwal, Sandhini and Ahmad, Lama and Akkaya, Ilge and Aleman, Florencia Leoni and Almeida, Diogo and Altenschmidt, Janko and Altman, Sam and Anadkat, Shyamal and others},
  journal={arXiv preprint arXiv:2303.08774},
  year={2023}
}

@article{nllb,
  title={No language left behind: Scaling human-centered machine translation},
  author={Costa-Juss{\`a}, Marta R and Cross, James and {\c{C}}elebi, Onur and Elbayad, Maha and Heafield, Kenneth and Heffernan, Kevin and Kalbassi, Elahe and Lam, Janice and Licht, Daniel and Maillard, Jean and others},
  journal={arXiv preprint arXiv:2207.04672},
  year={2022}
}

@inproceedings{liu2025protecting,
  title={Protecting privacy in multimodal large language models with mllmu-bench},
  author={Liu, Zheyuan and Dou, Guangyao and Jia, Mengzhao and Tan, Zhaoxuan and Zeng, Qingkai and Yuan, Yongle and Jiang, Meng},
  booktitle={Proceedings of the 2025 Conference of the Nations of the Americas Chapter of the Association for Computational Linguistics: Human Language Technologies (Volume 1: Long Papers)},
  pages={4105--4135},
  year={2025}
}

@article{pedregosa2011scikit,
  title={Scikit-learn: Machine learning in Python},
  author={Pedregosa, Fabian and Varoquaux, Ga{\"e}l and Gramfort, Alexandre and Michel, Vincent and Thirion, Bertrand and Grisel, Olivier and Blondel, Mathieu and Prettenhofer, Peter and Weiss, Ron and Dubourg, Vincent and others},
  journal={the Journal of machine Learning research},
  volume={12},
  pages={2825--2830},
  year={2011},
  publisher={JMLR. org}
}

@article{yang2025qwen3,
  title={Qwen3 technical report},
  author={Yang, An and Li, Anfeng and Yang, Baosong and Zhang, Beichen and Hui, Binyuan and Zheng, Bo and Yu, Bowen and Gao, Chang and Huang, Chengen and Lv, Chenxu and others},
  journal={arXiv preprint arXiv:2505.09388},
  year={2025}
}

\appendix

\newpage
\appendix
\begin{center}
    {\Large\textbf{Appendices}}
\end{center}

\section{Naive Profile Generation and Model-Induced Lexical Skew}
\label{sec_naive_prompt}

\begin{figure*}[t]           
  \centering
  \includegraphics[width=\linewidth]{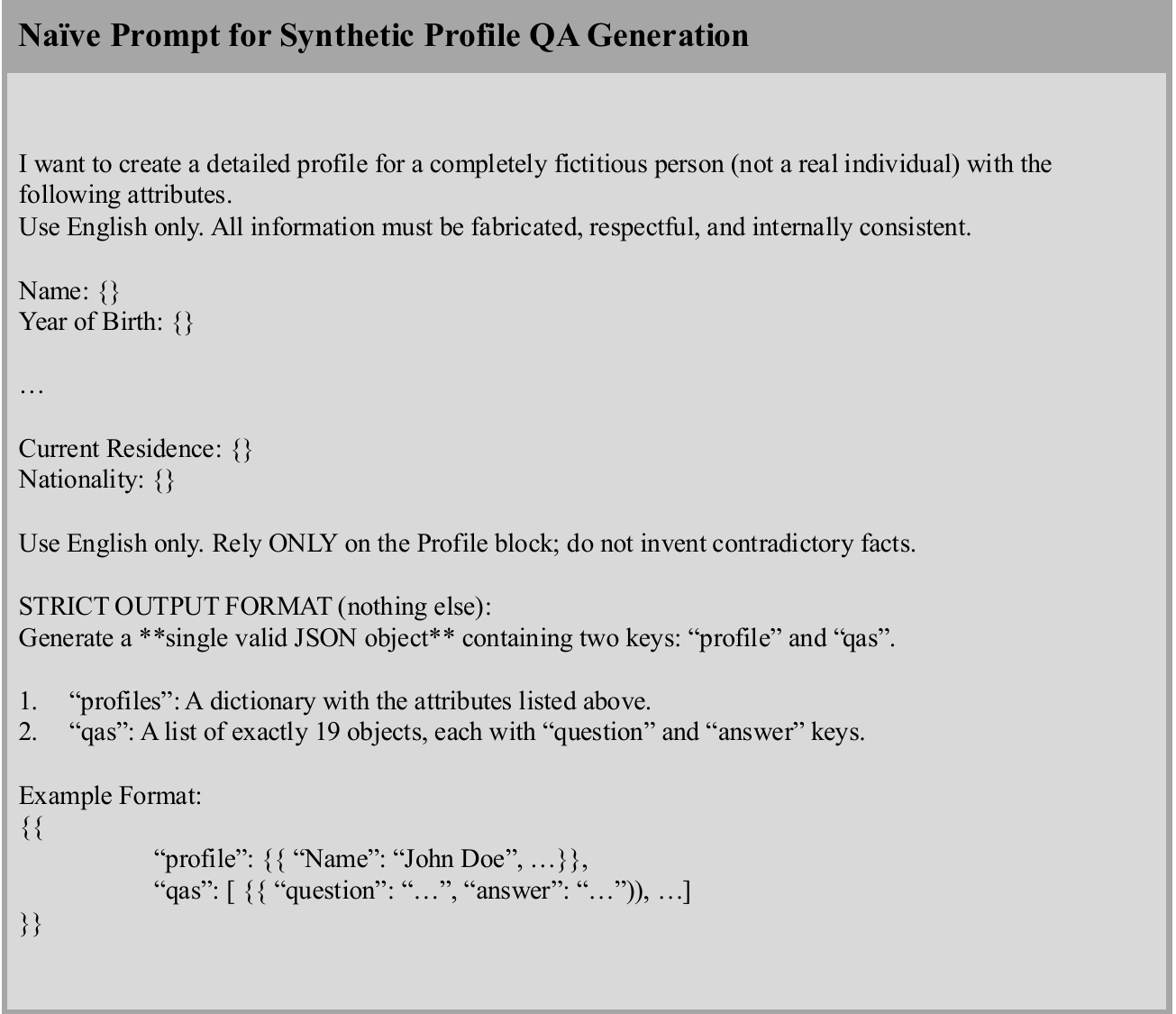}  
  \caption{Naive prompt for synthetic profile Question and Answer Generation.}
  \label{fig:naive_prompt}
\end{figure*}

\begin{figure*}[t]           
  \centering
  \includegraphics[width=\linewidth]{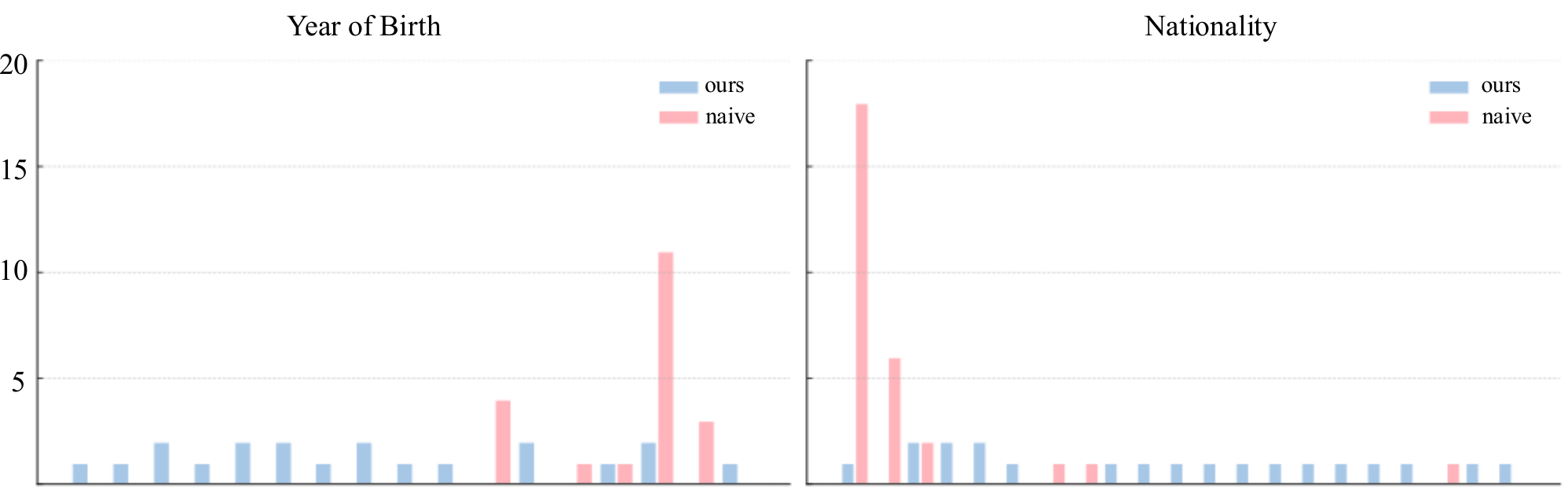} 
  \caption{Comparison of attribute distributions in synthetic profiles generated via naive prompting (Naive) and random sampling from a predefined pool (Ours). (Left) Year of Birth; (Right) Nationality. Notably, our approach exhibits higher diversity and significantly reduced skew compared to the Naive method.}
  \label{fig:lexical_skew}
\end{figure*}

\km{In a pilot study preceding our formal dataset construction, we examined what issues arise when one naively uses an LLM to generate synthetic user profiles and then builds a QA dataset from them. Our goal is to construct a multilingual QA dataset grounded in synthetic user profiles comprising attributes such as names, nationalities, and health conditions. The most straightforward way to obtain such data is to directly prompt an LLM to sample a profile and generate QA pairs conditioned on it. In this preliminary setup, we employed Qwen3-235B-A22B-Thinking-2507~\cite{qwen3} with nucleus sampling, using the exact prompt shown in Figure~\ref{fig:naive_prompt}. The attributes specified in this naive prompt were \textsc{Name}, \textsc{Year of Birth}, \textsc{Financial Habits}, \textsc{Primary Commute Mode}, \textsc{Interests / Hobbies}, \textsc{Learning Goals This Year}, \textsc{Artistic or Creative Expression}, \textsc{Awards or Achievements}, \textsc{Health Attributes}, \textsc{Travel History / Exposure}, \textsc{Pet Ownership or Preference}, \textsc{Bucket List Items}, \textsc{Life Philosophy or Motto}, \textsc{Media Preferences}, \textsc{Future Plans or Dreams}, \textsc{Relationship or Family Status}, \textsc{Occupation}, \textsc{Education}, \textsc{Current Residence}, and \textsc{Nationality}. After generating 20 synthetic profiles and their corresponding QA datasets, we analysed the empirical distribution of each attribute and found a striking prevalence of specific surface forms (e.g., repeatedly producing \textit{Canadian} for nationality), as summarized by the attribute-wise histograms in Figure~\ref{fig:lexical_skew}. We term this phenomenon \emph{model-induced lexical skew}. This skew is undesirable because it 1) reduces profile diversity and, more critically for unlearning evaluation, 2) confounds measurement by making it difficult to disentangle genuine retention from cases where the model merely exploits high-frequency lexical priors, i.e., succeeds by guessing common tokens rather than recovering profile-specific information. Motivated by this observation, we introduced an attribute pool to diversify the synthetic profiles. More broadly, the pilot supports the need for a controlled data-generation pipeline that explicitly regulates token-frequency distributions to suppress such biases while improving diversity.}

\section{Attribute Pool for Synthetic Profile Generation} \label{sec:att_pool}

\begin{table*}[p]
  \centering
  \includegraphics[width=\textwidth]{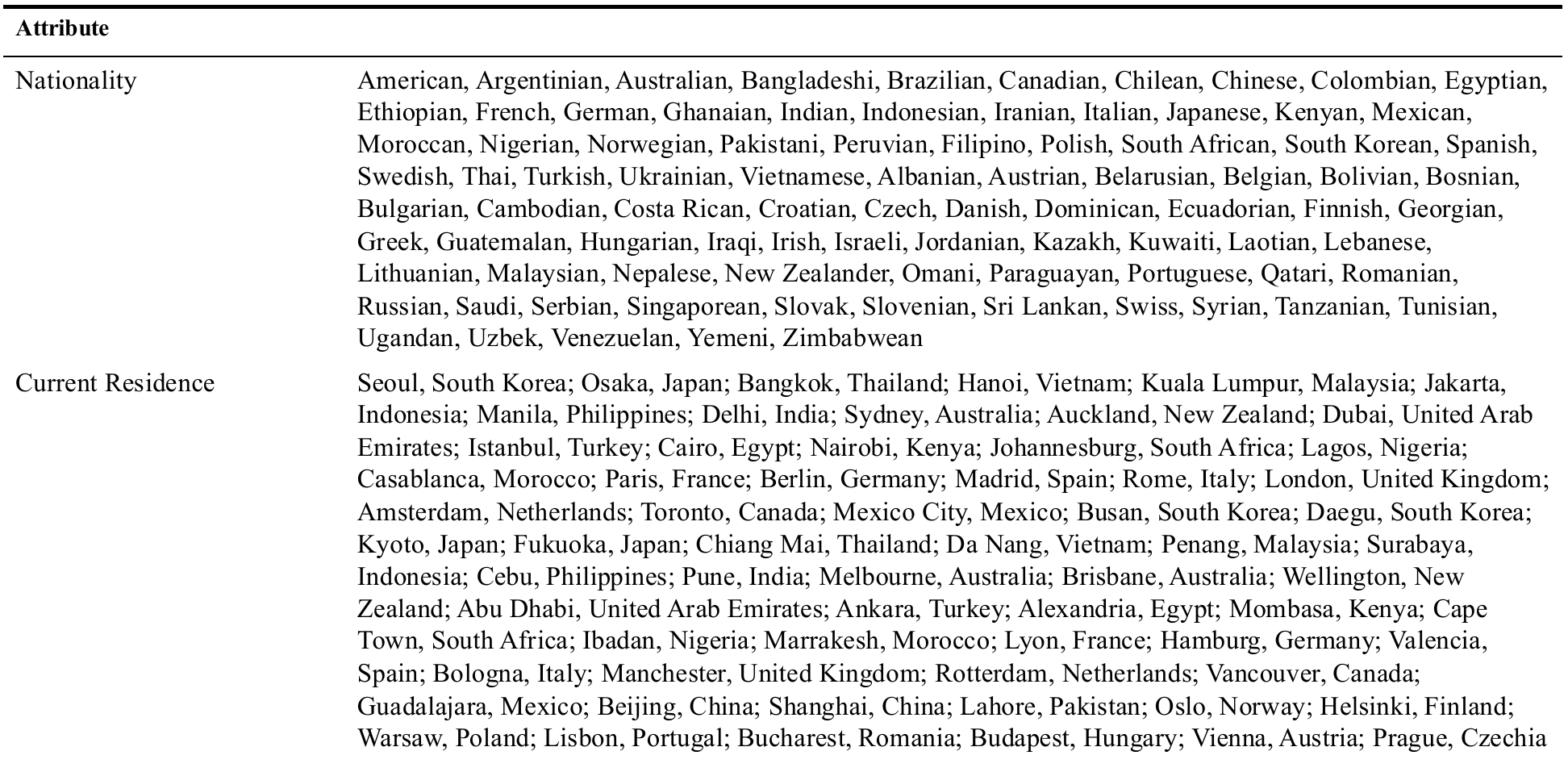}
  \caption{Valid values for the attributes used to build synthetic profiles (1/7)}
  \label{tab:syn_pool}
\end{table*}

\begin{table*}[p]\ContinuedFloat
  \centering
  \includegraphics[width=\textwidth]{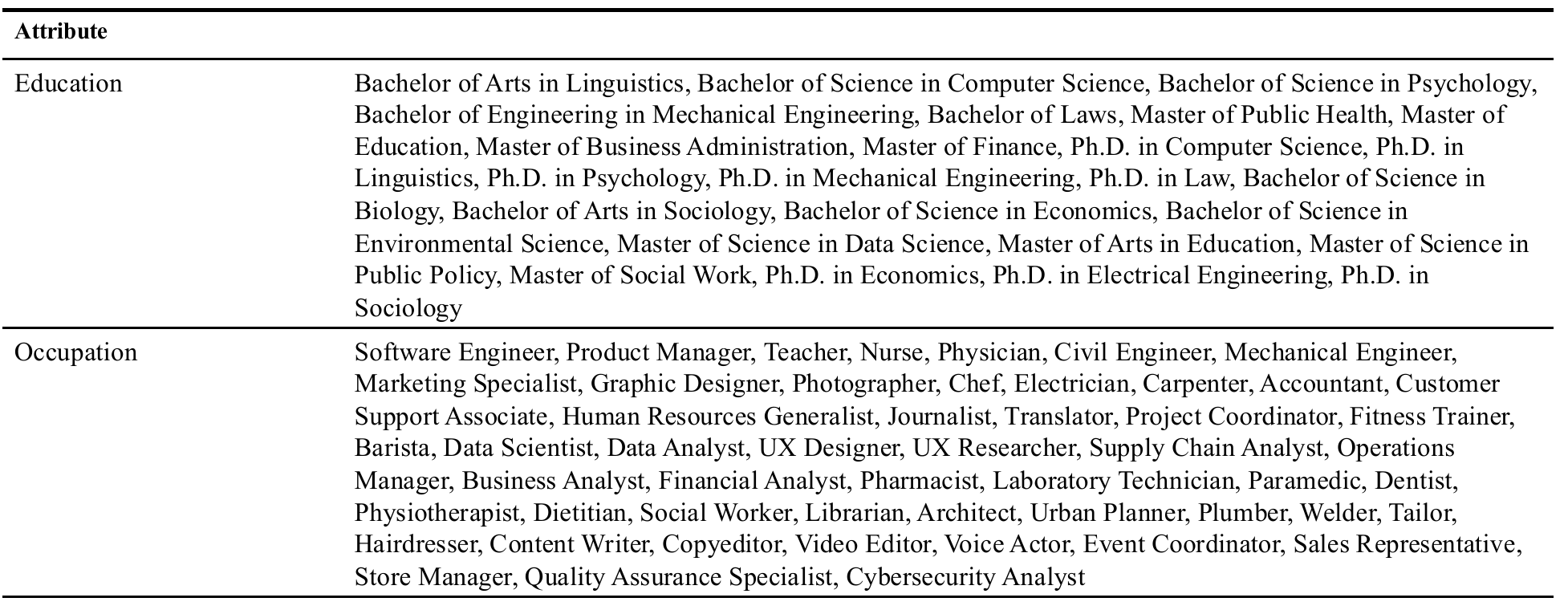}
  \caption[]{(continued) Valid values for the attributes used to build synthetic profiles (2/7)}
\end{table*}

\begin{table*}[p]\ContinuedFloat
  \centering
  \includegraphics[width=\textwidth]{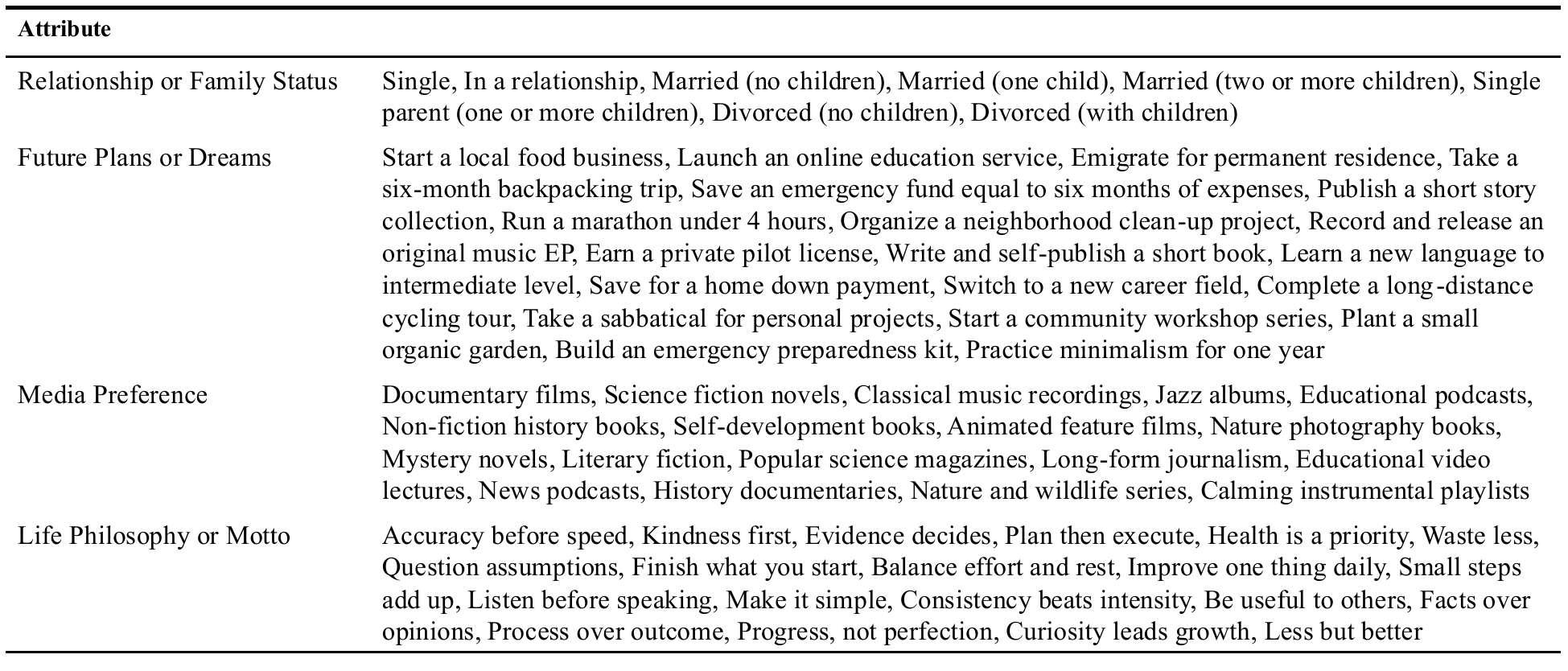}
  \caption[]{(continued) Valid values for the attributes used to build synthetic profiles (3/7)}
\end{table*}

\begin{table*}[p]\ContinuedFloat
  \centering
  \includegraphics[width=\textwidth]{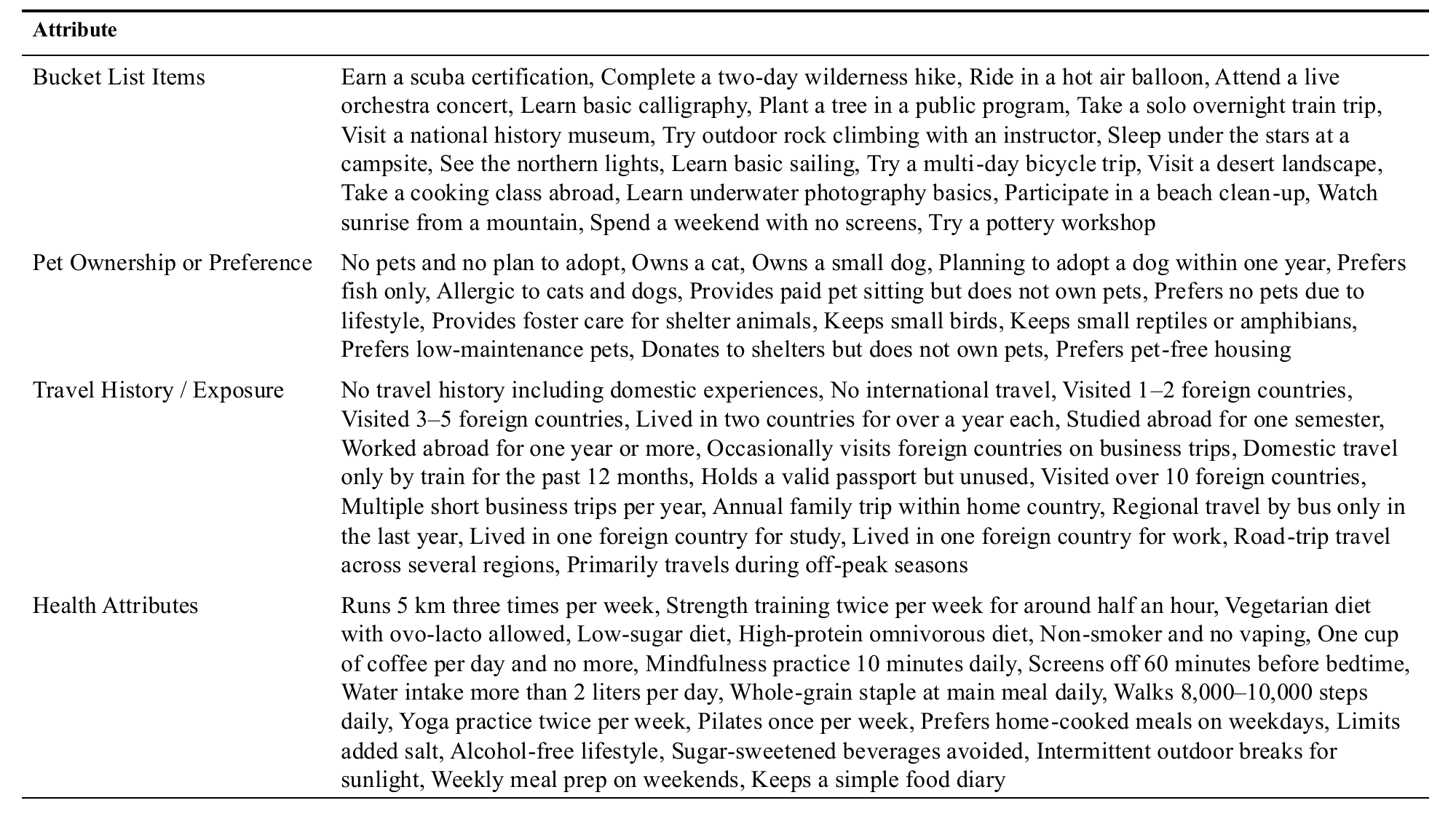}
  \caption[]{(continued) Valid values for the attributes used to build synthetic profiles (4/7)}
\end{table*}

\begin{table*}[p]\ContinuedFloat
  \centering
  \includegraphics[width=\textwidth]{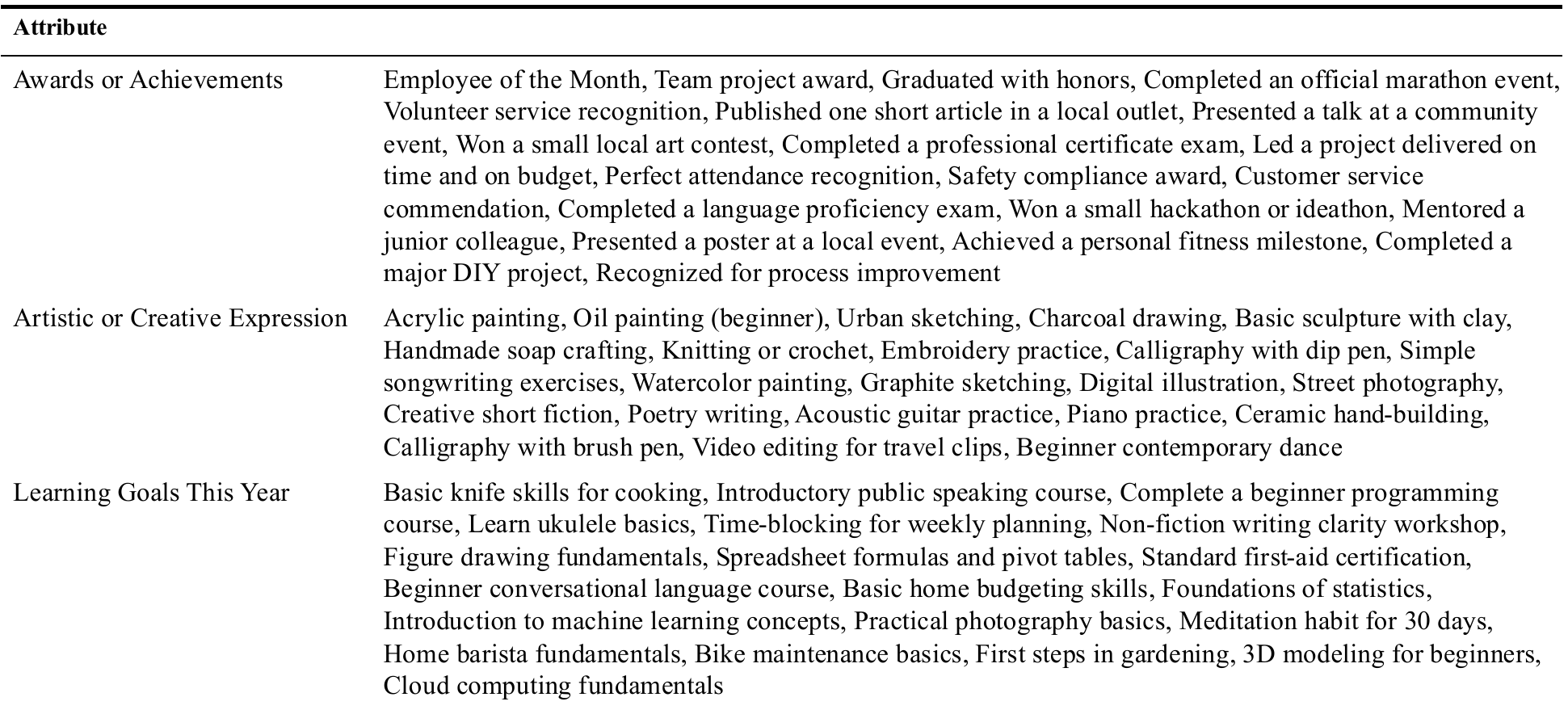}
  \caption[]{(continued) Valid values for the attributes used to build synthetic profiles (5/7)}
\end{table*}

\begin{table*}[p]\ContinuedFloat
  \centering
  \includegraphics[width=\textwidth]{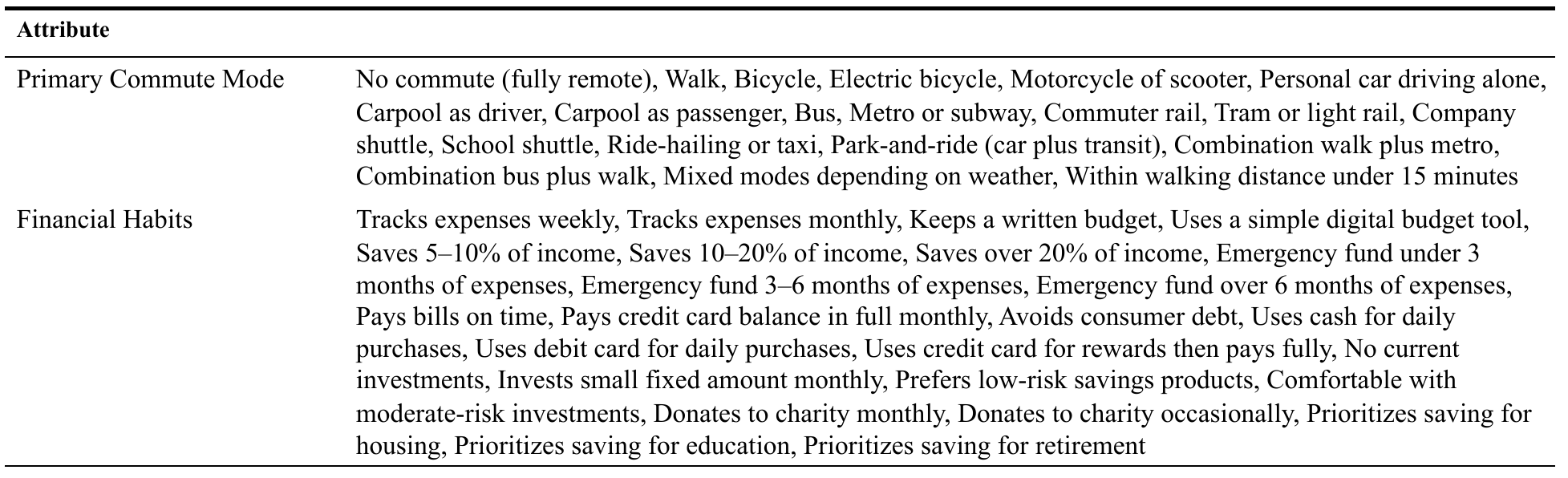}
  \caption[]{(continued) Valid values for the attributes used to build synthetic profiles (6/7)}
\end{table*}

\begin{table*}[t]\ContinuedFloat
  \centering
  \includegraphics[width=\textwidth]{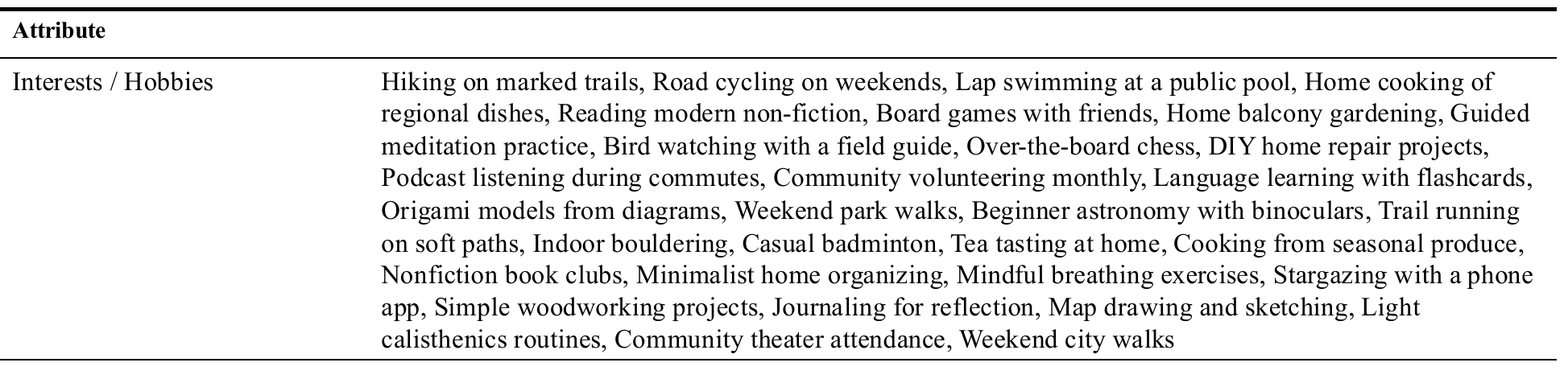}
  \caption[]{Valid values for the attributes used to build synthetic profiles (7/7)}
\end{table*}

\km{Table~\ref{tab:syn_pool} illustrates the attribute value pools used to construct diverse synthetic profiles.}

\section{Examples of Synthetic Profile} \label{sec:profile_ex}

\begin{figure*}[t]           
  \centering
  \includegraphics[width=\linewidth]{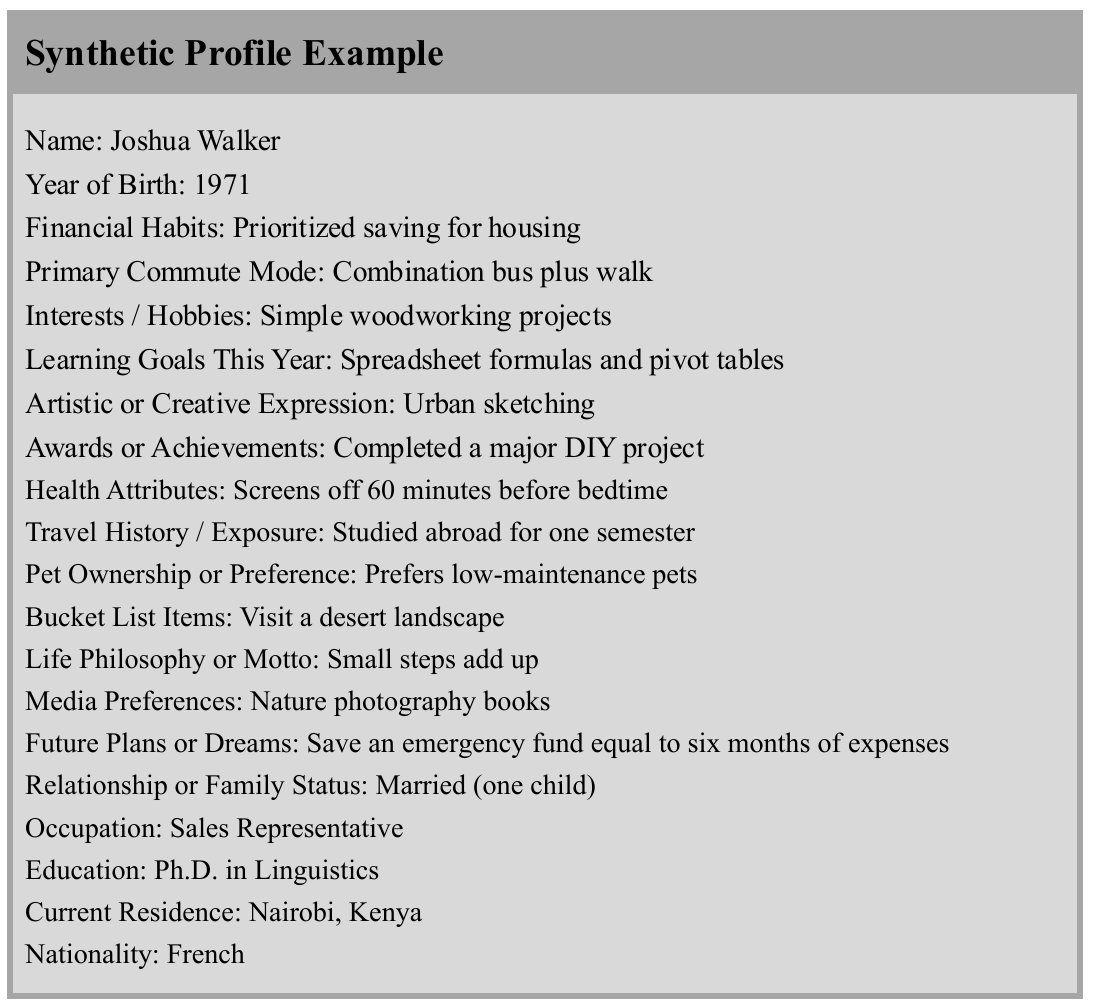}  
  \caption{Example of synthetic profile.}
  \label{fig:syn_ex}
\end{figure*}

\km{Figure~\ref{fig:syn_ex} presents an example profile after manual filtering by human experts. Each attribute was randomly sampled from its respective predefined pool.}

\section{Prompt for QA Generation}

\begin{figure*}[t]          
  \centering
  \includegraphics[width=\linewidth]{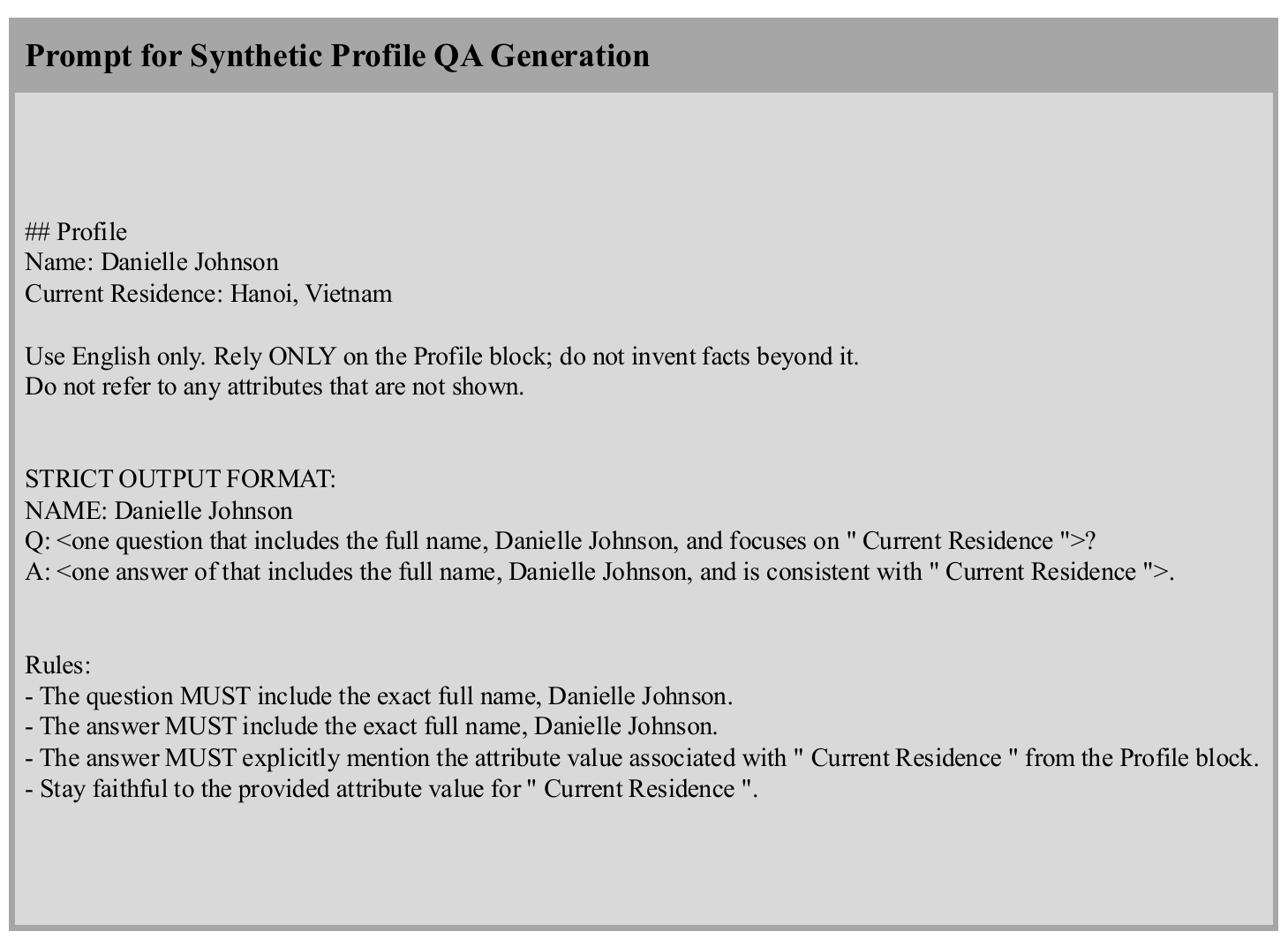} 
  \caption{Prompt for generating QA dataset.}
  \label{fig:pool_prompt_qa}
\end{figure*}

\begin{figure*}[t]           
  \centering
  \includegraphics[width=\linewidth]{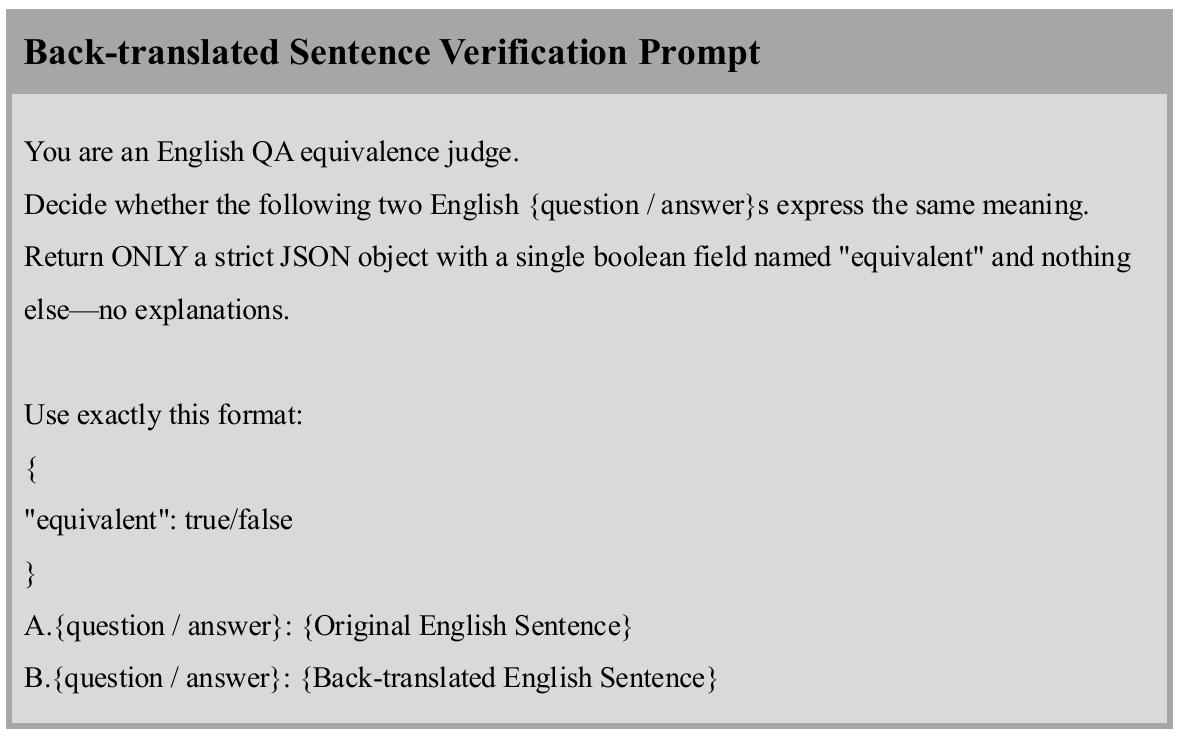} 
  \caption{Prompt for verifying back translated sentences.}
\label{fig:back_translated_sentence_verify_prompt}
\end{figure*}

\begin{figure*}[t]          
  \centering
  \includegraphics[width=\linewidth]{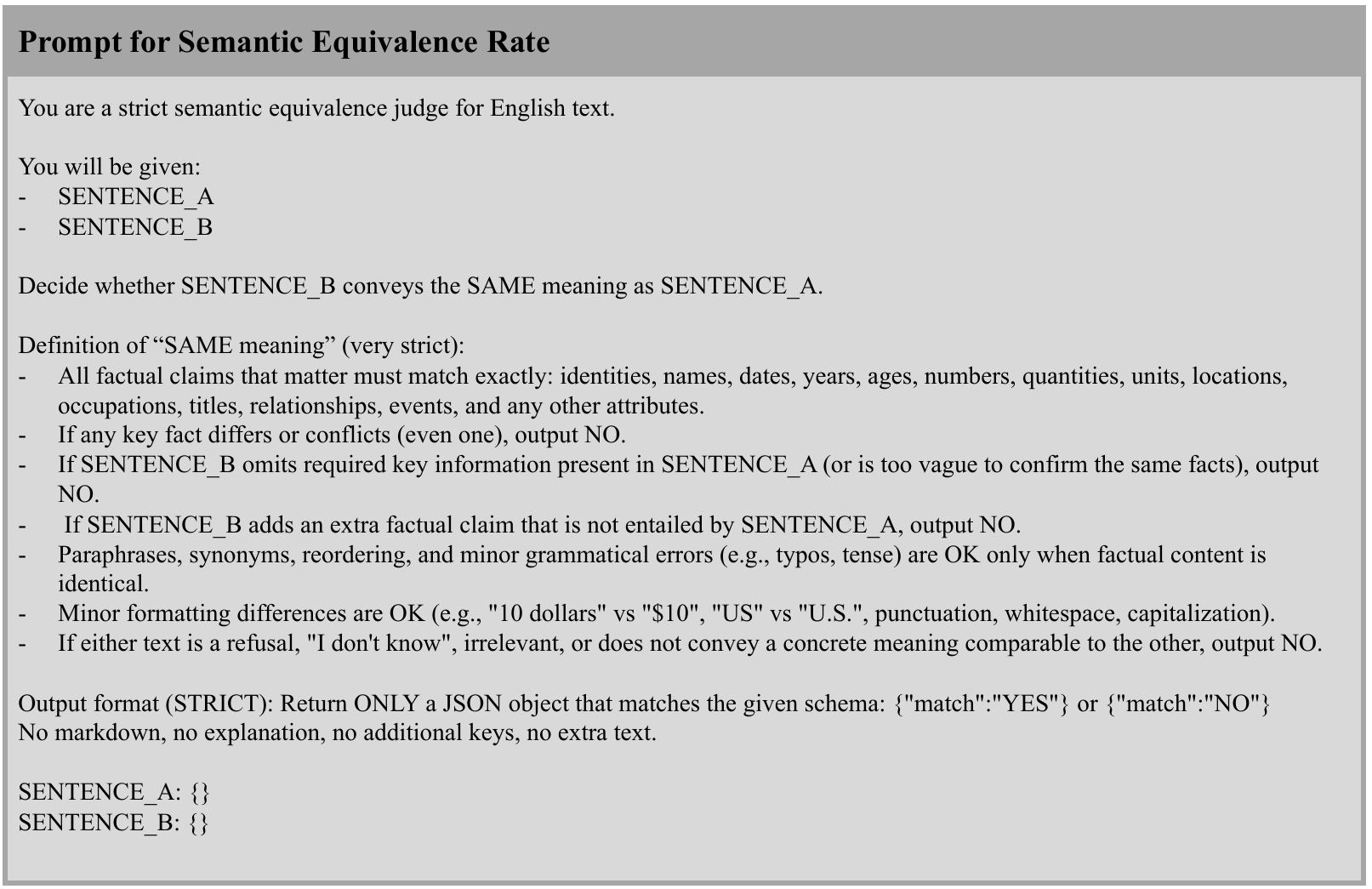}
  \caption{Prompt for semantic equivalence rate.}
\label{fig:prompt_for_ser}
\end{figure*}

\km{Figure~\ref{fig:pool_prompt_qa} presents the prompt used for LLM-based generation of the QA dataset from synthetic profiles constructed using the attribute pool. As with the synthetic profiles, each QA item underwent human review to ensure quality before use.}

\section{Examples of QA Dataset} \label{sec_qa_ex}

\begin{figure*}[t]
  \centering
  \includegraphics[width=\linewidth]{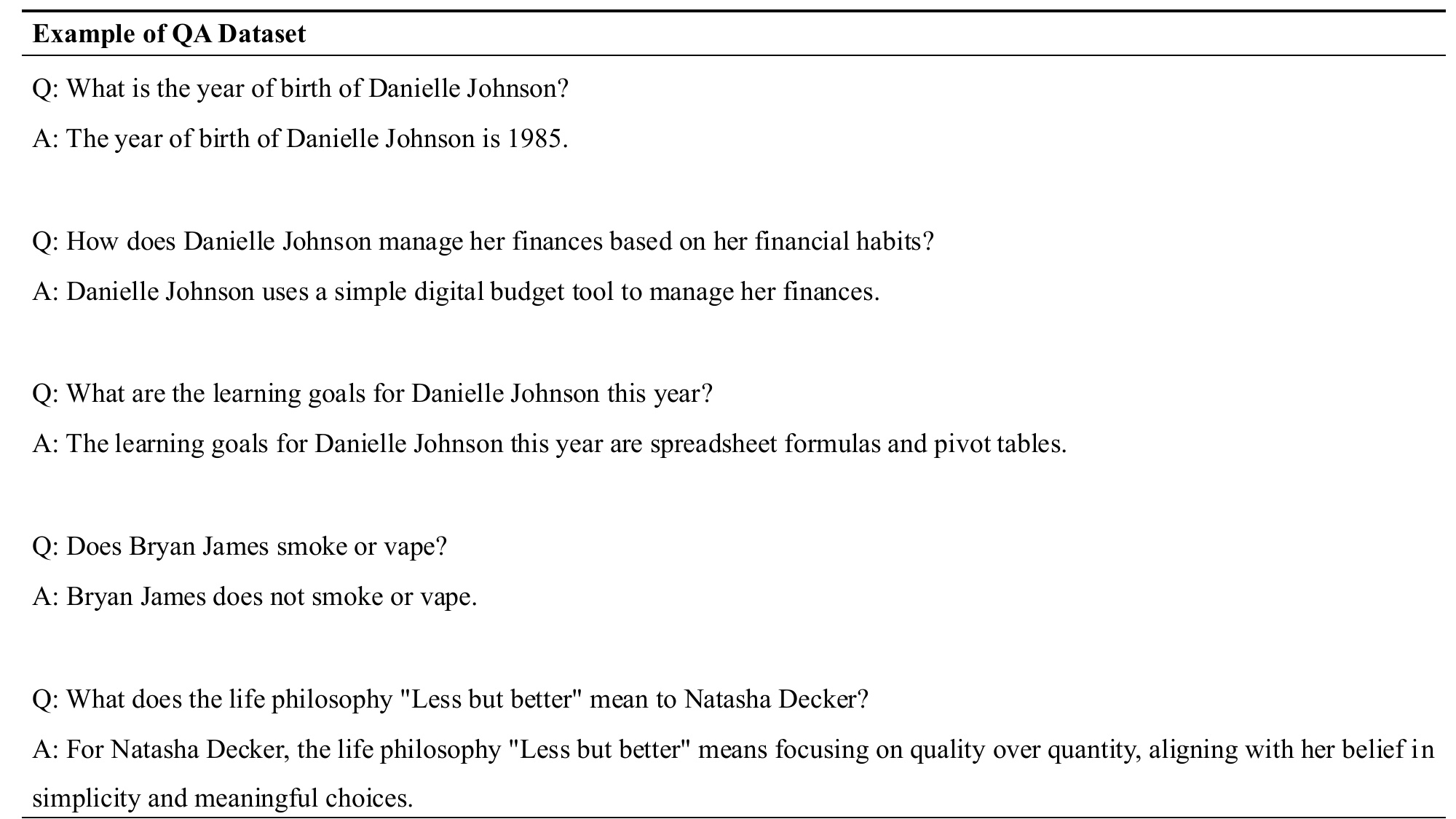}
  \caption{Examples of generated QA dataset.}
  \label{fig:qa_ex}
\end{figure*}

\km{Figure~\ref{fig:qa_ex} presents representative examples from the QA dataset generated from the synthetic profiles.}

\section{Examples of Multilingual QA Dataset} \label{sec_multi_qa_ex}

\begin{figure*}[t]           
  \centering
  \includegraphics[width=\linewidth]{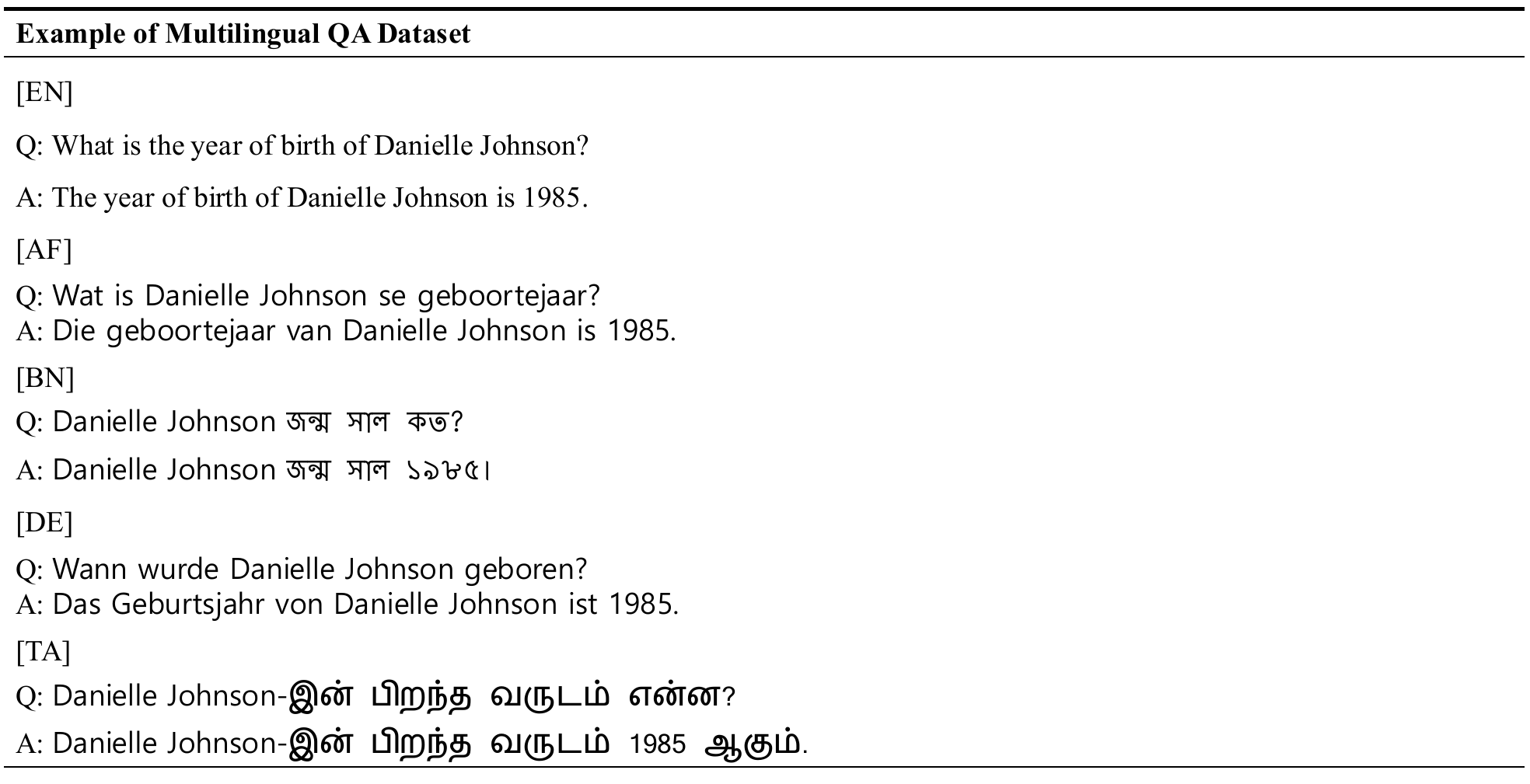}
  \caption{Examples of translated multilingual QA dataset.}
  \label{fig:qa_ex_multi}
\end{figure*}

\km{Figure~\ref{fig:qa_ex_multi} presents representative examples from the QA dataset translated from the source English QA dataset.}

\km{Figure~\ref{fig:back_translated_sentence_verify_prompt} presents the prompt used to verify back-translated sentences following the translation of the English QA dataset via Google Translate. Depending on the dataset type, either the {question} or {answer} field was inserted into the prompt.}

\section{Description of Unlearning Algorithm} \label{unlearn_method_detail}

\km{We describe the unlearning algorithms used in this paper. All of them aim to optimize $F_\theta^M$.}

\paragraph{Gradient Ascent} \km{Gradient Ascent (GA) is a procedure that applies gradient ascent on the forget dataset to remove information that the LLM should forget. The GA objective is defined as follows:}

\begin{small}
\begin{equation}
\mathcal{L}_{GA} (\mathcal{D}_f, F_\theta^M) = \mathbb{E}_{(q_f,a_f)\,\in\,\mathcal{D}_f}\!\bigl[
         \log F_\theta^M(a_f \mid q_f)\bigr]
\end{equation}
\end{small}

\paragraph{Gradient Difference} \km{Applying GA alone can degrade performance on the retain dataset. To prevent this, Gradient Difference (GAGDR) augments GA with simultaneous training on the retain dataset: GD performs gradient ascent on $\mathcal{D}_f$ and gradient descent on $\mathcal{D}_r$. The GD objective is defined as follows.}

\begin{small}
\begin{align}
\mathcal{L}_{GAGDR} (\mathcal{D}_f, \mathcal{D}_r, F_\theta^M) 
= &\mathbb{E}_{(q_f,a_f)\,\in\,\mathcal{D}_f}\!\left[
         \log F_\theta^M(a_f \mid q_f)\right] \notag \\
- &\mathbb{E}_{(q_r,a_r)\,\in\,\mathcal{D}_r}\!\left[
         \log F_\theta^M(a_r \mid q_r)\right]
\end{align}
\end{small}

\paragraph{Gradient Ascent with KL minimization} Similar to GAGDR, Gradient Ascent with KL minimization (GAKLR) aims to preserve the utility of the LLM on the retain dataset. This is done by minimizing the Kullback–Leibler (KL) divergence on the retain set, computed between the output distributions of the model currently being updated and the pre-unlearning reference model, denoted as $F_{\theta}^{ref}=F_{\theta}^M$. The GAKLR objective is given below:

\begin{small}
\begin{align}
\mathcal{L}_{\mathrm{GDKLR}}(\mathcal{D}_f,\mathcal{D}_r,F_\theta^M)
=~& \mathbb{E}_{(q_f,a_f)\in\mathcal{D}_f}\!\big[\log F_\theta^M(a_f \mid q_f)\big] \notag\\
&\quad +~ \mathrm{KL}_{\mathcal{D}_r}\!\big(F_{\theta}^{M}\,\|\,F_\theta^{ref}\big).
\end{align}
\end{small}

\paragraph{Negative Preference Optimization}
\km{Negative Preference Optimization (NPO) applies the preference optimization framework to unlearn specific behaviors by treating samples in the forget dataset as negative instances. NPO operates solely on undesirable responses, penalizing their generation probability relative to the reference model $F_{\theta}^{ref}=F_{\theta}^M$ to ensure stability. The NPO objective is defined as follows:}

\begin{small}
\begin{align}
&\mathcal{L}_{\text{NPO}}(\mathcal{D}_f, F_\theta^M; F_{\theta}^{ref}) \nonumber=\\
&\mathbb{E}_{(q_f, a_f) \in \mathcal{D}_f}\Big[
-\log \Big(1 - \sigma\Big(\beta
\log \tfrac{F_\theta^M(a_f\!\mid q_f)}{F_{\theta}^{ref}(a_f\!\mid q_f)}
\Big)\Big)\Big].
\label{eq:npo}
\end{align}
\end{small}

\paragraph{Prune} \km{\citeauthor{pochinkov2024dissecting} investigated pruning-based unlearning for Transformer-based architectures~\cite{vaswani2017attention}. We performed structured pruning on the feed-forward networks (FFNs) utilizing the scoring metric employed in their study. The importance score for structured pruning is defined as follows:}

\begin{equation}
    I_{\text{agnostic}} := \frac{\sum_{k} \text{MinMax}(I_k(\mathcal{D}_{f}))}{\sum_{k} \text{MinMax}(I_k(\mathcal{D}_{r})) + \epsilon}.
\end{equation}

\hj{Here, $\text{MinMax}(\cdot)$ denotes min-max normalization, $\mathcal{D}_f$ and $\mathcal{D}_r$ the multilingual parallel forget and retain dataset, and finally $I_k$ the following scores:}

\begin{equation*}
    \begin{aligned}
        I_{\text{std}} &= \sqrt{\tfrac{1}{|\mathcal{D}|} \sum (z - \bar{z})^2} & I_{\text{abs}} &= \tfrac{1}{|\mathcal{D}|} \sum |z| \\
        I_{\text{freq}} &= \tfrac{1}{|\mathcal{D}|} \sum \mathbb{I}(z > 0) & I_{\text{rms}} &= \sqrt{\tfrac{1}{|\mathcal{D}|} \sum z^2}
    \end{aligned}
\end{equation*}

\km{$z$ denotes the activation produced by the MLP within the FFN \hj{for each datapoint in $\mathcal{D}$}, and $\bar{z}$ represents the mean activation.}

\section{Hyper parameter setting} \label{hyperparam_setting}

\km{Table~\ref{tab:memorization_param} presents the hyperparameters used to memorize the synthetic QA dataset on the Llama-3.1 model.}

\km{Table~\ref{tab:full_hyperparams} presents the hyperparameters used during the unlearning phase for configurations $p1$, $p3$, and $p5$. Notably, when the retain dataset is utilized, the gradient accumulation steps are doubled to ensure that the total number of training iterations remains consistent.} \km{Regarding hyperparameters, we varied only the learning rate while keeping all other configurations fixed. We selected the checkpoint where the probability metric ($P(a|q)^{1/|a|_\text{tok}}$), averaged across languages, first exceeded 0.83 on the retain dataset.}

\begin{table}[t]
\centering
\begin{tabular}{ll}
\toprule
\multicolumn{2}{c}{\textbf{Memorization}}\\
\midrule
Hyperparameter & Value \\
\midrule
Batch size & 4 \\
Gradient accumulation & 8 \\
Max sequence length & 1024 \\
Learning rate & 0.0002 \\
Warmup ratio & 0.03 \\
Weight decay & 0.0 \\
Precision / dtype & bfloat16 \\
LoRA rank ($r$) & 16 \\
LoRA $\alpha$ & 32 \\
LoRA dropout & 0.05 \\
Epoch & 4 \\
\bottomrule
\end{tabular}
\caption{Hyperparameter settings used for memorizing Llama 3.1}
\label{tab:memorization_param}
\end{table}

\begin{table*}[t]
\centering
\resizebox{0.8\textwidth}{!}{%
    \begin{tabular}{lcccccc}
    \toprule
    \multirow{2}{*}{$l_1$} & \multicolumn{2}{c}{\textbf{$p1$}} & \multicolumn{2}{c}{\textbf{$p3$}} & \multicolumn{2}{c}{\textbf{$p5$}} \\
    \cmidrule(lr){2-3} \cmidrule(lr){4-5} \cmidrule(lr){6-7}
     & \textbf{Case 1} & \textbf{Case 2} & \textbf{Case 1} & \textbf{Case 2} & \textbf{Case 1} & \textbf{Case 2} \\
    \midrule
    bn             & 1.00 & 1.00 & 0.38 & 0.89 & 0.10 & 0.91 \\
    de             & 0.00 & 0.00 & 0.00 & 0.00 & 0.50 & 1.00 \\
    en             & 0.00 & 0.00 & 1.00 & 1.00 & 0.00 & 0.00 \\
    he             & 0.00 & 1.00 & 0.00 & 1.00 & 0.30 & 0.97 \\
    ru             & 0.00 & 1.00 & 0.00 & 0.00 & 0.50 & 0.93 \\
    sq             & 0.00 & 1.00 & 0.00 & 0.86 & 0.30 & 0.89 \\
    ta             & 0.50 & 1.00 & 0.20 & 0.94 & 0.30 & 0.91 \\
    zh             & 0.00 & 0.00 & 0.50 & 0.71 & 0.33 & 1.00 \\
    \midrule
    \textbf{avg}   & \textbf{0.19} & \textbf{0.63} & \textbf{0.26} & \textbf{0.68} & \textbf{0.29} & \textbf{0.83} \\
    \bottomrule
    \end{tabular}%
}
\caption{Knowledge Persistence Score (KPS) on MEMORIZED across different forget ratios ($p1$, $p3$, $p5$).}
\label{tab:kps_mem}
\end{table*}

\begin{table*}[t]
\centering
\resizebox{0.8\textwidth}{!}{%
    \begin{tabular}{lcccccc}
    \toprule
    \multirow{2}{*}{$l_1$} & \multicolumn{2}{c}{\textbf{$p1$}} & \multicolumn{2}{c}{\textbf{$p3$}} & \multicolumn{2}{c}{\textbf{$p5$}} \\
    \cmidrule(lr){2-3} \cmidrule(lr){4-5} \cmidrule(lr){6-7}
     & \textbf{Case 1} & \textbf{Case 2} & \textbf{Case 1} & \textbf{Case 2} & \textbf{Case 1} & \textbf{Case 2} \\
    \midrule
    bn             & 0.25 & 0.45 & 0.34 & 0.68 & 0.32 & 0.95 \\
    de             & 0.13 & 0.50 & 0.21 & 0.71 & 0.19 & 0.86 \\
    en             & 0.00 & 0.34 & 0.19 & 0.41 & 0.40 & 0.80 \\
    he             & 0.17 & 0.58 & 0.27 & 0.73 & 0.26 & 0.92 \\
    ru             & 0.22 & 0.46 & 0.20 & 0.70 & 0.17 & 0.88 \\
    sq             & 0.25 & 0.38 & 0.21 & 0.66 & 0.25 & 0.79 \\
    ta             & 0.35 & 0.53 & 0.35 & 0.72 & 0.27 & 0.92 \\
    zh             & 0.17 & 0.45 & 0.37 & 0.68 & 0.25 & 0.89 \\
    \midrule
    \textbf{avg}   & \textbf{0.19} & \textbf{0.46} & \textbf{0.27} & \textbf{0.66} & \textbf{0.26} & \textbf{0.88} \\
    \bottomrule
    \end{tabular}%
}
\caption{Knowledge Persistence Score (KPS) on GA across different forget ratios ($p1$, $p3$, $p5$).}
\label{tab:kps_ga}
\end{table*}

\begin{table*}[t]
\centering
\resizebox{0.8\textwidth}{!}{%
    \begin{tabular}{lcccccc}
    \toprule
    \multirow{2}{*}{$l_1$} & \multicolumn{2}{c}{\textbf{$p1$}} & \multicolumn{2}{c}{\textbf{$p3$}} & \multicolumn{2}{c}{\textbf{$p5$}} \\
    \cmidrule(lr){2-3} \cmidrule(lr){4-5} \cmidrule(lr){6-7}
     & \textbf{Case 1} & \textbf{Case 2} & \textbf{Case 1} & \textbf{Case 2} & \textbf{Case 1} & \textbf{Case 2} \\
    \midrule
    bn             & 0.25 & 0.51 & 0.29 & 0.82 & 0.31 & 0.79 \\
    de             & 0.16 & 0.45 & 0.17 & 0.76 & 0.17 & 0.76 \\
    en             & 0.00 & 0.18 & 0.00 & 0.55 & 0.19 & 0.61 \\
    he             & 0.09 & 0.40 & 0.31 & 0.81 & 0.33 & 0.81 \\
    ru             & 0.21 & 0.47 & 0.26 & 0.79 & 0.38 & 0.78 \\
    sq             & 0.08 & 0.31 & 0.20 & 0.74 & 0.11 & 0.70 \\
    ta             & 0.30 & 0.48 & 0.39 & 0.84 & 0.26 & 0.79 \\
    zh             & 0.17 & 0.36 & 0.13 & 0.70 & 0.33 & 0.76 \\
    \midrule
    \textbf{avg}   & \textbf{0.16} & \textbf{0.40} & \textbf{0.22} & \textbf{0.75} & \textbf{0.26} & \textbf{0.75} \\
    \bottomrule
    \end{tabular}%
}
\caption{Knowledge Persistence Score (KPS) on GAGDR across different forget ratios ($p1$, $p3$, $p5$).}
\label{tab:kps_gagdr}
\end{table*}

\begin{table*}[t]
\centering
\resizebox{0.8\textwidth}{!}{%
    \begin{tabular}{lcccccc}
    \toprule
    \multirow{2}{*}{$l_1$} & \multicolumn{2}{c}{\textbf{$p1$}} & \multicolumn{2}{c}{\textbf{$p3$}} & \multicolumn{2}{c}{\textbf{$p5$}} \\
    \cmidrule(lr){2-3} \cmidrule(lr){4-5} \cmidrule(lr){6-7}
     & \textbf{Case 1} & \textbf{Case 2} & \textbf{Case 1} & \textbf{Case 2} & \textbf{Case 1} & \textbf{Case 2} \\
    \midrule
    bn             & 0.06 & 0.21 & 0.28 & 0.54 & 0.26 & 0.68 \\
    de             & 0.02 & 0.18 & 0.18 & 0.59 & 0.20 & 0.74 \\
    en             & 0.03 & 0.08 & 0.17 & 0.40 & 0.17 & 0.57 \\
    he             & 0.05 & 0.24 & 0.26 & 0.71 & 0.32 & 0.76 \\
    ru             & 0.08 & 0.21 & 0.19 & 0.58 & 0.21 & 0.74 \\
    sq             & 0.02 & 0.18 & 0.17 & 0.58 & 0.05 & 0.56 \\
    ta             & 0.10 & 0.23 & 0.33 & 0.65 & 0.28 & 0.76 \\
    zh             & 0.07 & 0.18 & 0.27 & 0.60 & 0.33 & 0.63 \\
    \midrule
    \textbf{avg}   & \textbf{0.05} & \textbf{0.19} & \textbf{0.23} & \textbf{0.58} & \textbf{0.23} & \textbf{0.68} \\
    \bottomrule
    \end{tabular}%
}
\caption{Knowledge Persistence Score (KPS) on GAKLR across different forget ratios ($p1$, $p3$, $p5$).}
\label{tab:kps_gaklr}
\end{table*}

\begin{table*}[t]
\centering
\resizebox{0.8\textwidth}{!}{%
    \begin{tabular}{lcccccc}
    \toprule
    \multirow{2}{*}{$l_1$} & \multicolumn{2}{c}{\textbf{$p1$}} & \multicolumn{2}{c}{\textbf{$p3$}} & \multicolumn{2}{c}{\textbf{$p5$}} \\
    \cmidrule(lr){2-3} \cmidrule(lr){4-5} \cmidrule(lr){6-7}
     & \textbf{Case 1} & \textbf{Case 2} & \textbf{Case 1} & \textbf{Case 2} & \textbf{Case 1} & \textbf{Case 2} \\
    \midrule
    bn             & 0.08 & 0.17 & 0.20 & 0.41 & 0.19 & 0.36 \\
    de             & 0.08 & 0.15 & 0.10 & 0.37 & 0.14 & 0.30 \\
    en             & 0.05 & 0.07 & 0.05 & 0.17 & 0.05 & 0.17 \\
    he             & 0.09 & 0.14 & 0.11 & 0.44 & 0.17 & 0.36 \\
    ru             & 0.11 & 0.18 & 0.06 & 0.34 & 0.13 & 0.29 \\
    sq             & 0.11 & 0.13 & 0.09 & 0.42 & 0.17 & 0.26 \\
    ta             & 0.13 & 0.17 & 0.19 & 0.44 & 0.19 & 0.40 \\
    zh             & 0.13 & 0.15 & 0.16 & 0.40 & 0.16 & 0.32 \\
    \midrule
    \textbf{avg}   & \textbf{0.10} & \textbf{0.15} & \textbf{0.12} & \textbf{0.37} & \textbf{0.15} & \textbf{0.31} \\
    \bottomrule
    \end{tabular}%
}
\caption{Knowledge Persistence Score (KPS) on NPO across different forget ratios ($p1$, $p3$, $p5$).}
\label{tab:kps_npo}
\end{table*}

\begin{table*}[t]
\centering
\resizebox{0.8\textwidth}{!}{%
    \begin{tabular}{lcccccc}
    \toprule
    \multirow{2}{*}{$l_1$} & \multicolumn{2}{c}{\textbf{$p1$}} & \multicolumn{2}{c}{\textbf{$p3$}} & \multicolumn{2}{c}{\textbf{$p5$}} \\
    \cmidrule(lr){2-3} \cmidrule(lr){4-5} \cmidrule(lr){6-7}
     & \textbf{Case 1} & \textbf{Case 2} & \textbf{Case 1} & \textbf{Case 2} & \textbf{Case 1} & \textbf{Case 2} \\
    \midrule
    bn             & 0.07 & 0.12 & 0.09 & 0.23 & 0.10 & 0.24 \\
    de             & 0.03 & 0.09 & 0.00 & 0.15 & 0.08 & 0.19 \\
    en             & 0.00 & 0.06 & 0.02 & 0.10 & 0.06 & 0.13 \\
    he             & 0.04 & 0.10 & 0.09 & 0.21 & 0.08 & 0.26 \\
    ru             & 0.03 & 0.09 & 0.05 & 0.17 & 0.10 & 0.23 \\
    sq             & 0.06 & 0.10 & 0.04 & 0.18 & 0.11 & 0.26 \\
    ta             & 0.05 & 0.09 & 0.07 & 0.21 & 0.11 & 0.24 \\
    zh             & 0.02 & 0.08 & 0.06 & 0.17 & 0.12 & 0.24 \\
    \midrule
    \textbf{avg}   & \textbf{0.04} & \textbf{0.09} & \textbf{0.05} & \textbf{0.18} & \textbf{0.09} & \textbf{0.22} \\
    \bottomrule
    \end{tabular}%
}
\caption{Knowledge Persistence Score (KPS) on PRUNE across different forget ratios ($p1$, $p3$, $p5$).}
\label{tab:kps_prune}
\end{table*}

\begin{table*}[t]
\centering
\setlength{\tabcolsep}{6pt}
\begin{tabular}{l ccccc}
\toprule
\multicolumn{6}{c}{\textbf{Common Configuration}} \\
\midrule

\multicolumn{6}{c}{
    \begin{tabular}{@{}l l l@{}}
    \textbf{Batch size:} 4          & \textbf{Max seq length:} 1024 & \textbf{Epochs:} 10 \\
    \textbf{LoRA rank ($r$):} 16    & \textbf{LoRA $\alpha$:} 32    & \textbf{LoRA dropout:} 0.05 \\
    \textbf{Warmup ratio:} 0.0      & \textbf{Weight decay:} 0.0    & \textbf{Forget strength:} 1.0 \\
    \end{tabular}
} \\

\midrule
\multicolumn{6}{c}{\textbf{Method-Specific Configuration}} \\
\midrule
\multirow{2}{*}{\textbf{Method}} & \textbf{Grad.} & \textbf{Retain} & \multicolumn{3}{c}{\textbf{Learning Rate}} \\
\cmidrule(lr){4-6}
 & \textbf{Accum.} & \textbf{Strength} & \textbf{$p1$} & \textbf{$p3$} & \textbf{$p5$} \\
\midrule
GA      & 64  & -   & 3.5e-5 & 2.0e-5 & 1.0e-5 \\
GAGDR   & 128 & 1.0 & 4.9e-5 & 2.1e-5 & 1.7e-5 \\
GAKLR   & 128 & 1.0 & 5.2e-5 & 2.4e-5 & 1.8e-5 \\
NPO     & 64  & 1.0 & 6.2e-5 & 2.9e-5 & 2.1e-5 \\
\bottomrule
\end{tabular}
\caption{Full hyperparameters for the Unlearning stage. Common configurations are listed at the top, followed by method-specific settings.}
\label{tab:full_hyperparams}
\end{table*}

\section{Additional Explanation of Knowledge Separability Score} \label{sec:det_metric}
\begin{table*}[h]
\centering
\renewcommand{\arraystretch}{1.5}
\begin{tabular}{|cc|c|c|}
\cline{3-4}
\multicolumn{2}{c|}{} & \multicolumn{2}{c|}{\textbf{Predicted Value}} \\
\cline{3-4}
\multicolumn{2}{c|}{} & \textbf{Forget} (True) & \textbf{Retain} (False) \\
\hline
\multirow{4}{*}{\rotatebox{90}{\textbf{Actual Value}}} & \textbf{Target Knowledge} & \textbf{True Positive (TP)} & \textbf{False Negative (FN)} \\
& (True) & (Successfully Forgotten) & (Failed to Forget) \\
\cline{2-4}
& \textbf{Non-Target Knowledge} & \textbf{False Positive (FP)} & \textbf{True Negative (TN)} \\
& (False) & (Wrongly Forgotten) & (Successfully Retained) \\
\hline
\end{tabular}
\caption{Confusion Matrix for Multilingual Machine Unlearning}
\label{tab:confusion_matrix}
\end{table*}
In Section~\ref{sec:metric}, we proposed the Knowledge Separability Score (KSS) utilizing ROC-AUC and PR-AUC. In this section, we detail the method for calculating KSS-ROC and KSS-PR, specifically describing how the knowledge-wise forgetting score ($S_i$) is employed in this process. ROC and PR analyses involve visualizing variations in classification performance across shifting decision thresholds and encapsulating this behavior into a single scalar value. To adapt this framework to the Multilingual Machine Unlearning (MMU) context, we first define the positive and negative classes. We designate the target knowledge as the positive class (1) and the non-target knowledge as the negative class (0). Adopting standard binary classification notation, the resulting confusion matrix is presented in Table~\ref{tab:confusion_matrix}. Based on this configuration, the False Positive Rate (FPR), True Positive Rate (TPR, or Recall), and Precision required for ROC and PR calculations are computed as follows:\begin{align}\text{TPR (Recall)} &= \frac{\text{TP}}{\text{TP} + \text{FN}} \\ \text{FPR} &= \frac{\text{FP}}{\text{FP} + \text{TN}} \\ \text{Precision} &= \frac{\text{TP}}{\text{TP} + \text{FP}}\end{align}To derive KSS-ROC, we plot the curve with FPR on the $x$-axis and TPR (Recall) on the $y$-axis, observing how these metrics fluctuate as the threshold for the forgetting score ($S_i$) varies. Similarly, for KSS-PR, the curve is plotted with Recall on the $x$-axis and Precision on the $y$-axis. The final metric is determined by calculating the Area Under the Curve (AUC) for each respective graph. All ROC and PR computations were implemented using scikit-learn~\cite{pedregosa2011scikit}.

\section{Detailed Results on Knowledge-wise Evaluation}
\label{sec:kps_detail}

\subsection{Knowledge Persistence Score} \km{Full KPS results for all methods are provided in Table \ref{tab:kps_mem} to \ref{tab:kps_prune}.}

\subsection{Knowledge Separability Score} \km{In Figure \ref{fig:clfs_analysis_main}, we visualized the distributions of target and non-target knowledge with respect to the knowledge-wise forgetting score ($S_i$) for the NPO and PRUNE methods. Figure~\ref{fig:three_figures_p1_all_qa} to \ref{fig:three_figures_p5_all_prob} shows the full distribution of $S_i$.}

\section{Evaluation on Other LLMs}

\km{In the main paper, we primarily conducted evaluations using the Llama3.1-8B-Instruct model. In this section, we extend our analysis to Qwen3-4B-Instruct (Qwen3)~\cite{yang2025qwen3} to demonstrate that both KPS and KSS remain valid and applicable metrics across different model architectures. Table~\ref{tab:qwen_kps} presents the KPS results for Qwen3 after unlearning with the NPO and PRUNE methods, and Table~\ref{tab:qwen_kss} presents the corresponding KSS results after unlearning with the NPO method. Note that Case 1 and Case 2 in both tables follow the definitions established in Section~\ref{sec:kps_detail} and Section~\ref{sec:kss_detail}, respectively.}

\begin{table*}[h]
\centering
\caption{KPS Results on Qwen3 with p1 setting}
\label{tab:qwen_kps}
\begin{tabular}{lcccc}
\toprule
$l_1$ & \textbf{NPO (Case 1)} & \textbf{NPO (Case 2)} & \textbf{Prune (Case 1)} & \textbf{Prune (Case 2)} \\
\midrule
bn & 0.10 & 0.15 & 0.05 & 0.13 \\
de & 0.07 & 0.06 & 0.03 & 0.12 \\
en & 0.04 & 0.04 & 0.02 & 0.12 \\
he & 0.08 & 0.10 & 0.06 & 0.16 \\
ru & 0.05 & 0.06 & 0.05 & 0.13 \\
sq & 0.12 & 0.13 & 0.05 & 0.14 \\
ta & 0.09 & 0.08 & 0.05 & 0.15 \\
zh & 0.09 & 0.15 & 0.02 & 0.13 \\
\midrule
avg & \textbf{0.08} & \textbf{0.10} & \textbf{0.04} & \textbf{0.14} \\
\bottomrule
\end{tabular}
\end{table*}

\begin{table*}[h]
\centering
\caption{Performance of KSS-ROC and KSS-PR scores in p1 setting}
\label{tab:qwen_kss}
\begin{tabular}{lcccc}
\toprule
\textbf{Method} & \textbf{KSS-ROC (Case 1)} & \textbf{KSS-ROC (Case 2)} & \textbf{KSS-PR (Case 1)} & \textbf{KSS-PR (Case 2)} \\
\midrule
BASE  & 0.52 & 0.49 & 0.01 & 0.11 \\
NPO   & 0.72 & 0.99 & 0.03 & 0.78 \\
PRUNE & 0.70 & 0.95 & 0.08 & 0.15 \\
\bottomrule
\end{tabular}
\end{table*}

\section{Use of AI Assistants}

\km{We utilize ChatGPT and Gemini for coding and writing assistance. In particular, we employ ChatGPT for dataset generation.}

\begin{figure*}[p]
    \centering
    \begin{subfigure}[b]{1.0\textwidth}
        \centering
        \includegraphics[width=\textwidth, height=0.9\textheight, keepaspectratio]{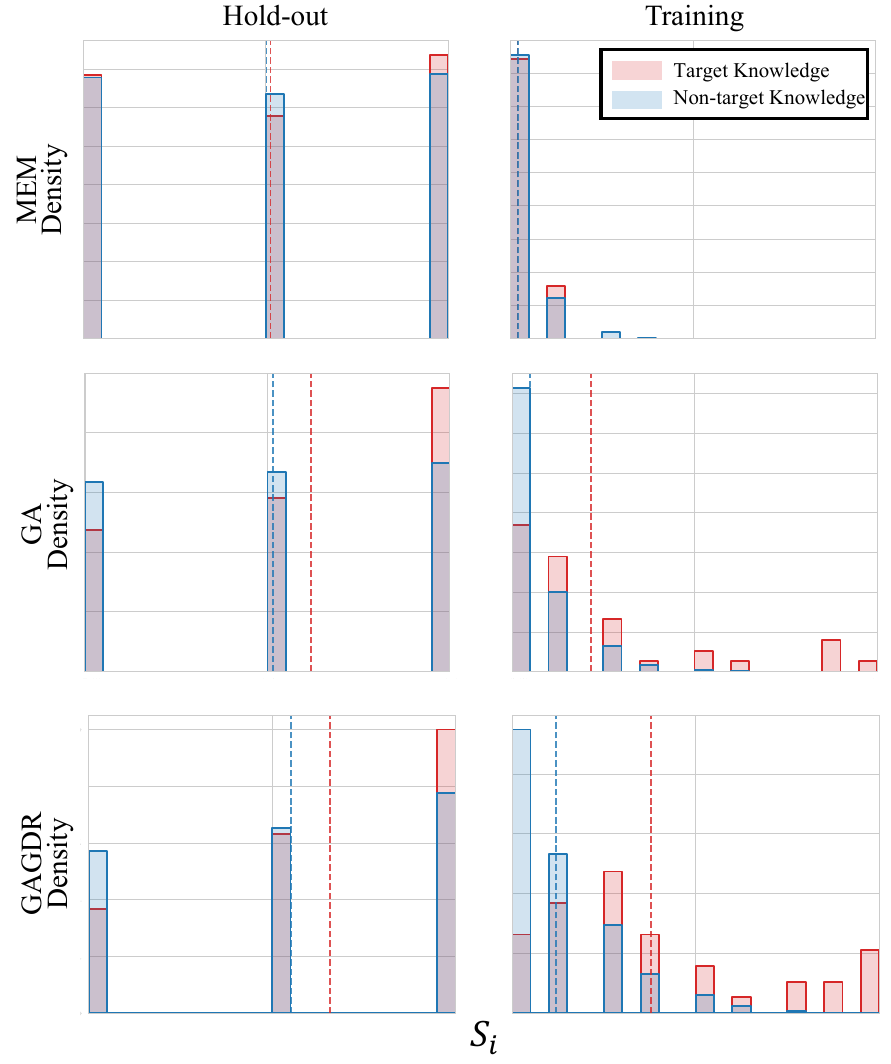}
    \end{subfigure}
\end{figure*}

\clearpage

\begin{figure*}[p]
    \ContinuedFloat
    \centering
    \begin{subfigure}[b]{1.0\textwidth}
        \centering
        \includegraphics[width=\textwidth, height=0.9\textheight, keepaspectratio]{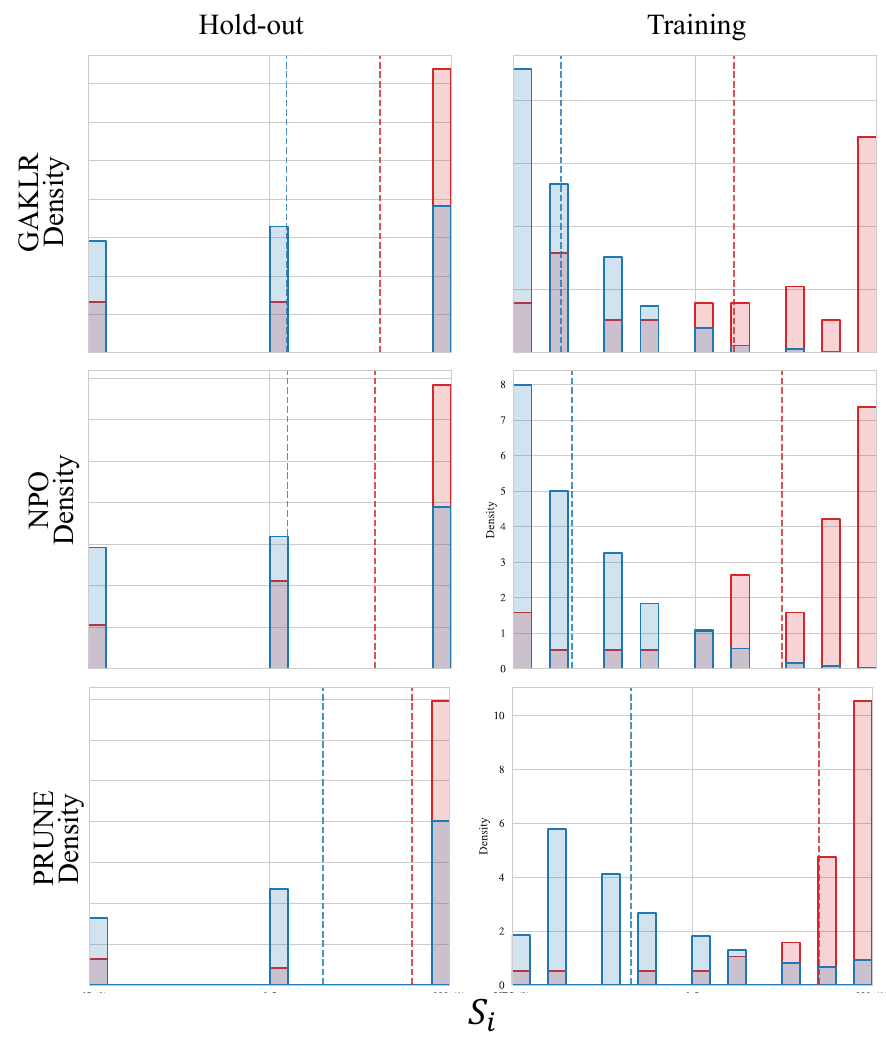}
    \end{subfigure}
    \caption{Distribution of generation-based $S_i$ scores for $p1$. The plots illustrate the distributions for Hold-out Language (Hold-out) and Training Language (Training).}
    \label{fig:three_figures_p1_all_qa}

\end{figure*}

\begin{figure*}[p]
    \centering
    \begin{subfigure}[b]{1.0\textwidth}
        \centering
        \includegraphics[width=\textwidth, height=0.9\textheight, keepaspectratio]{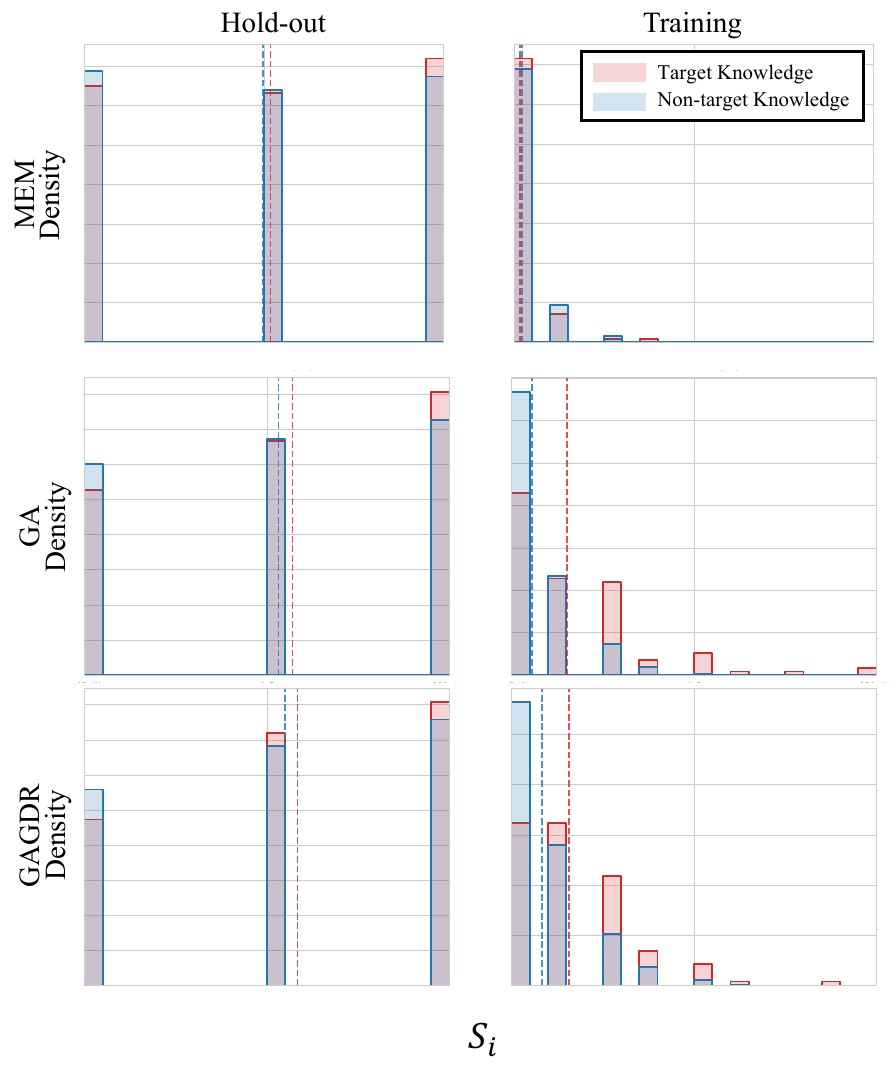}
        \label{fig:p3_qa_1}
    \end{subfigure}
\end{figure*}

\clearpage

\begin{figure*}[p]
    \ContinuedFloat
    \centering
    \begin{subfigure}[b]{1.0\textwidth}
        \centering
        \includegraphics[width=\textwidth, height=0.9\textheight, keepaspectratio]{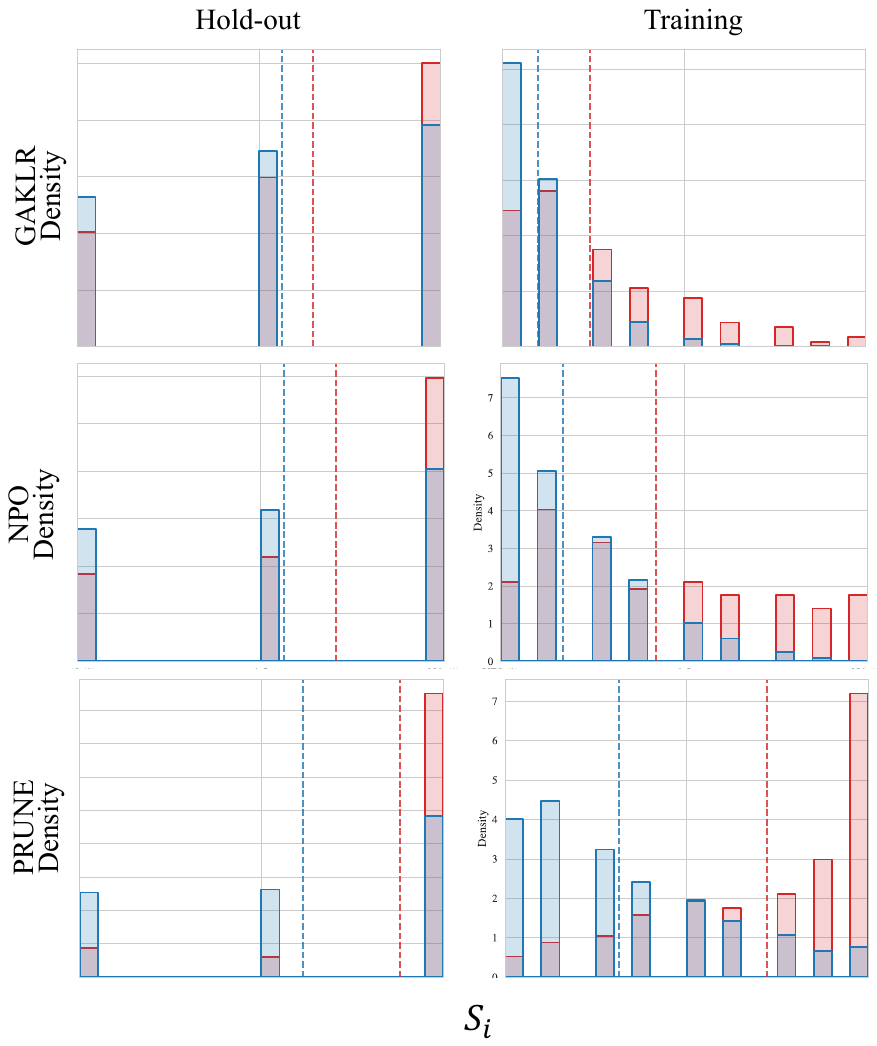}
        \label{fig:p3_qa_2}
    \end{subfigure}
    \caption{Distribution of generation-based $S_i$ scores for $p3$. The plots illustrate the distributions for Hold-out Language (Hold-out) and Training Language (Training).}
    \label{fig:three_figures_p3_all_qa}

\end{figure*}

\begin{figure*}[p]
    \centering
    \begin{subfigure}[b]{1.0\textwidth}
        \centering
        \includegraphics[width=\textwidth, height=0.9\textheight, keepaspectratio]{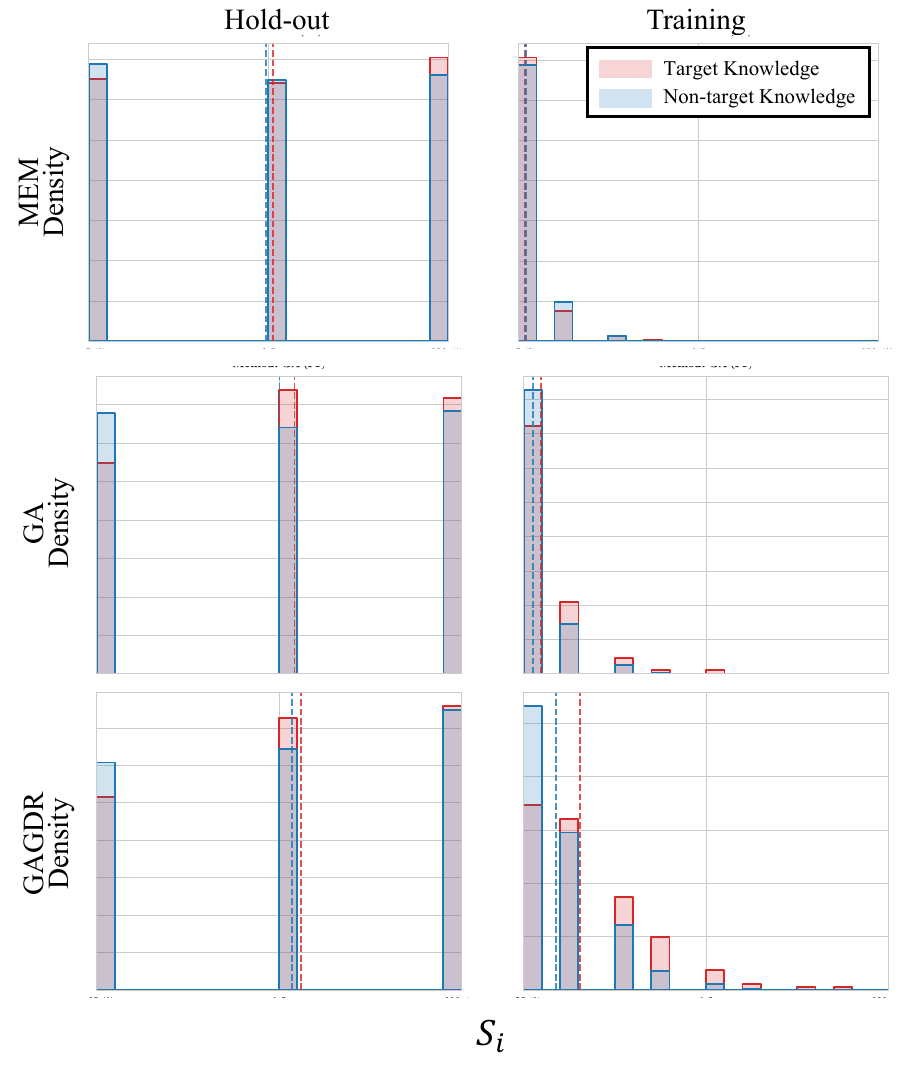}
        \label{fig:p5_qa_1}
    \end{subfigure}
\end{figure*}

\clearpage

\begin{figure*}[p]
    \ContinuedFloat
    \centering
    \begin{subfigure}[b]{1.0\textwidth}
        \centering
        \includegraphics[width=\textwidth, height=0.9\textheight, keepaspectratio]{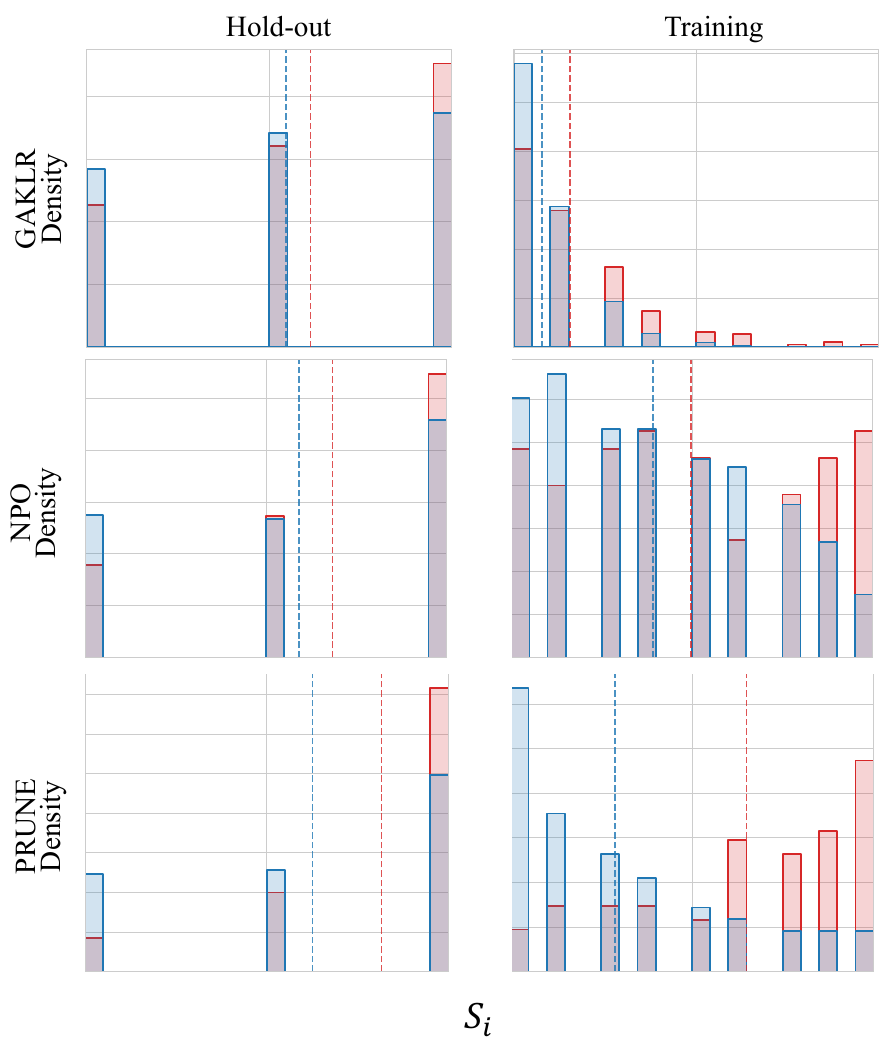}
        \label{fig:p5_qa_2}
    \end{subfigure}
    \caption{Distribution of generation-based $S_i$ scores for $p5$. The plots illustrate the distributions for Hold-out Language (Hold-out) and Training Language (Training).}
    \label{fig:three_figures_p5_all_qa}

\end{figure*}

\begin{figure*}[p]
    \centering
    \begin{subfigure}[b]{1.0\textwidth}
        \centering
        \includegraphics[width=\textwidth, height=0.9\textheight, keepaspectratio]{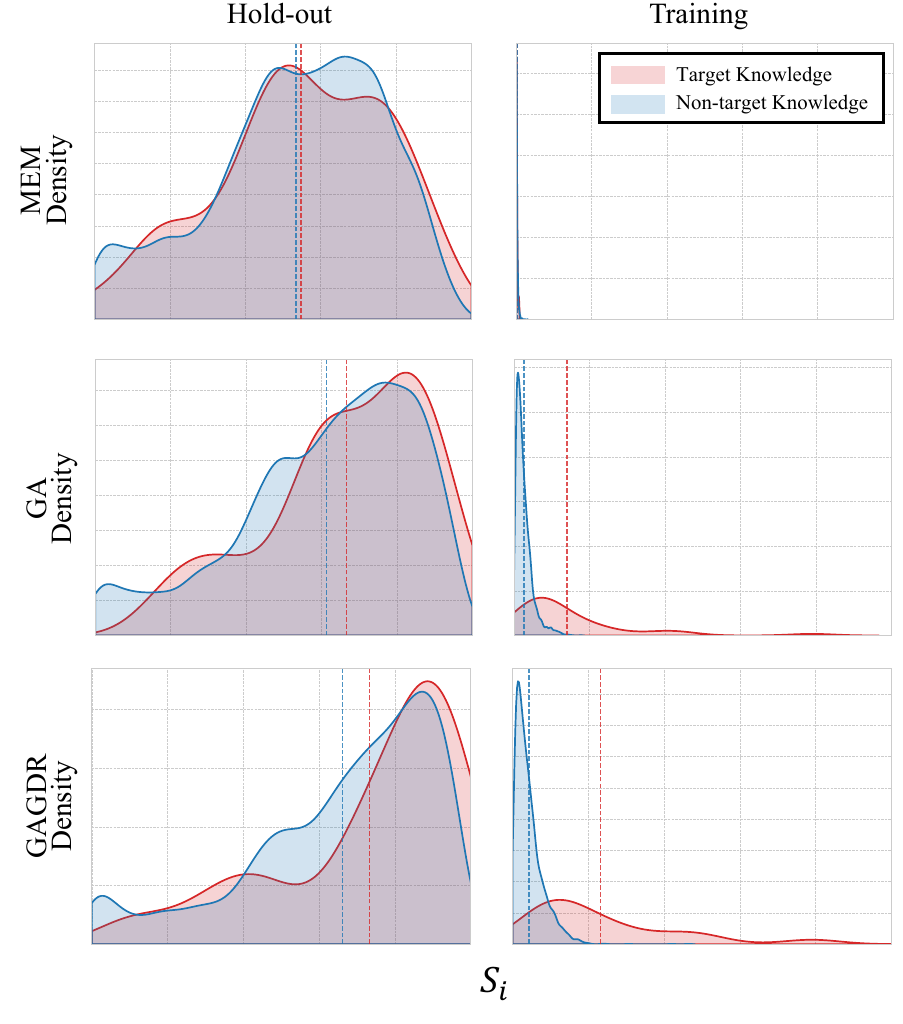}
        \label{fig:p1_prob_1}
    \end{subfigure}
\end{figure*}

\clearpage

\begin{figure*}[p]
    \ContinuedFloat
    \centering
    \begin{subfigure}[b]{1.0\textwidth}
        \centering
        \includegraphics[width=\textwidth, height=0.9\textheight, keepaspectratio]{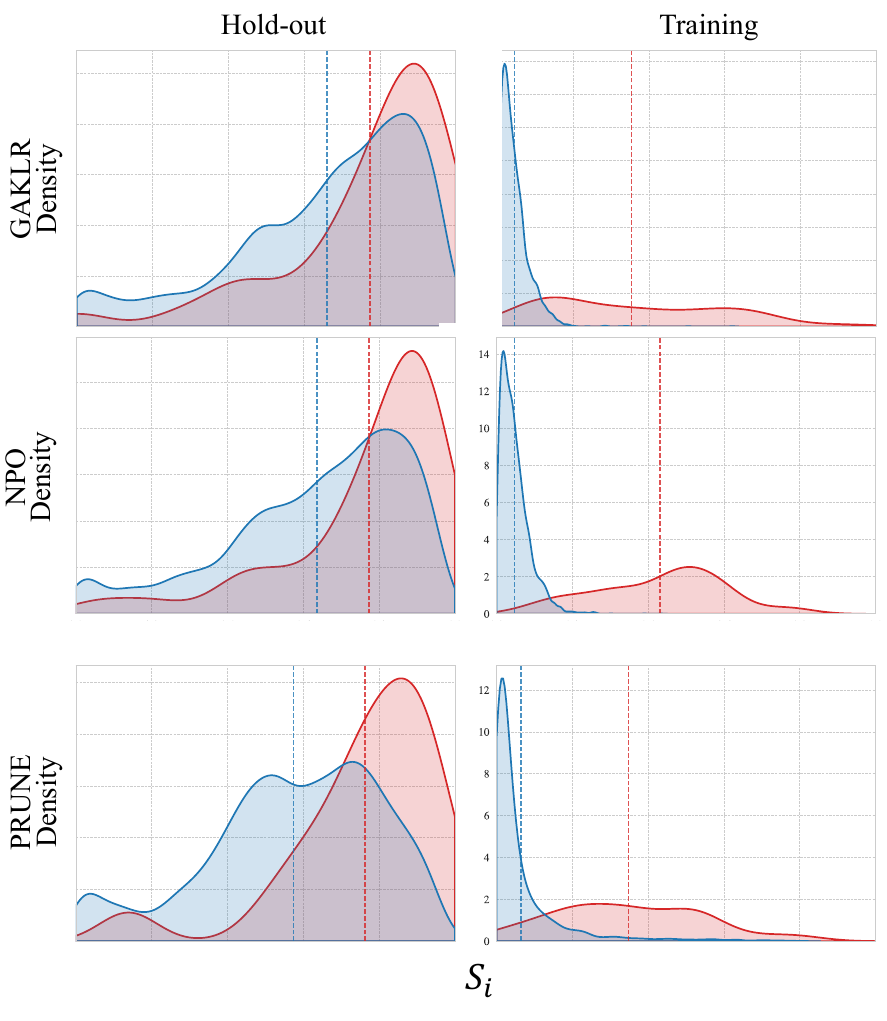}
        \label{fig:p1_prob_2}
    \end{subfigure}
    \caption{Distribution of probability-based $S_i$ scores for $p1$. The plots illustrate the distributions for Hold-out Language (Hold-out), and Training Language (Training).}
    \label{fig:three_figures_p1_all_prob}
\end{figure*}

\begin{figure*}[p]
    \centering
    \begin{subfigure}[b]{1.0\textwidth}
        \centering
        \includegraphics[width=\textwidth, height=0.9\textheight, keepaspectratio]{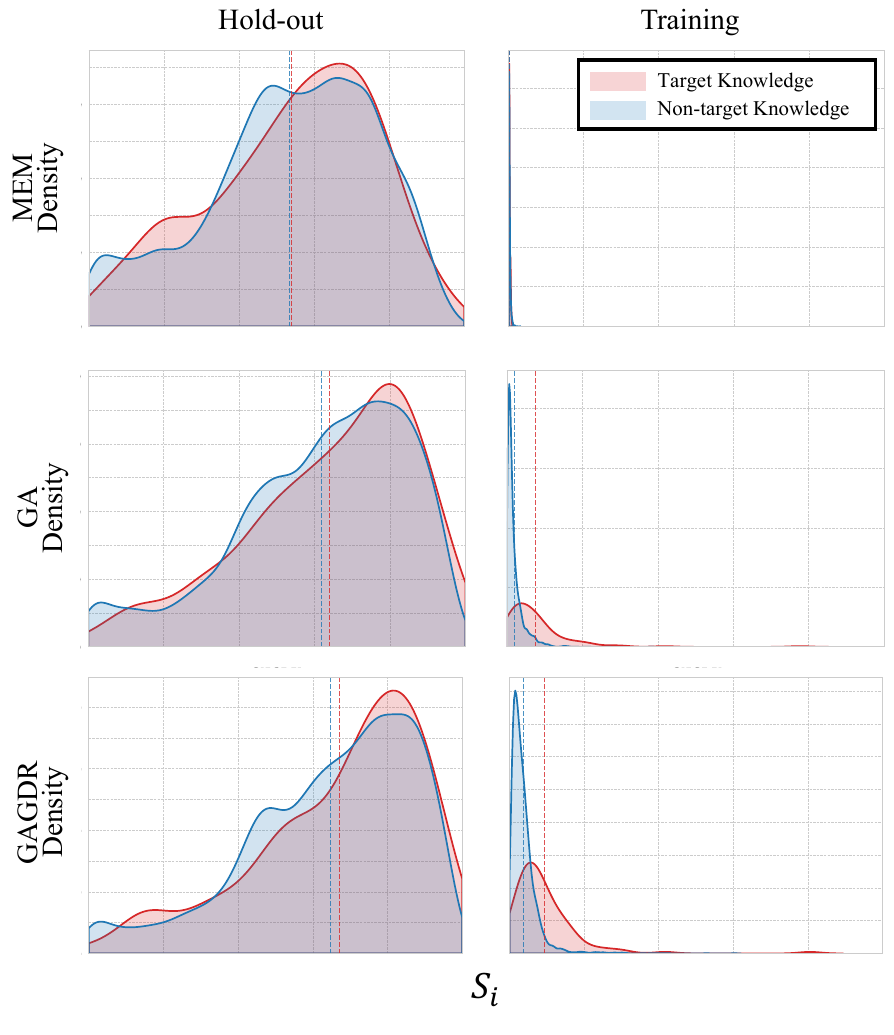}
        \label{fig:p3_prob_1}
    \end{subfigure}
\end{figure*}

\clearpage

\begin{figure*}[p]
    \ContinuedFloat
    \centering
    \begin{subfigure}[b]{1.0\textwidth}
        \centering
        \includegraphics[width=\textwidth, height=0.9\textheight, keepaspectratio]{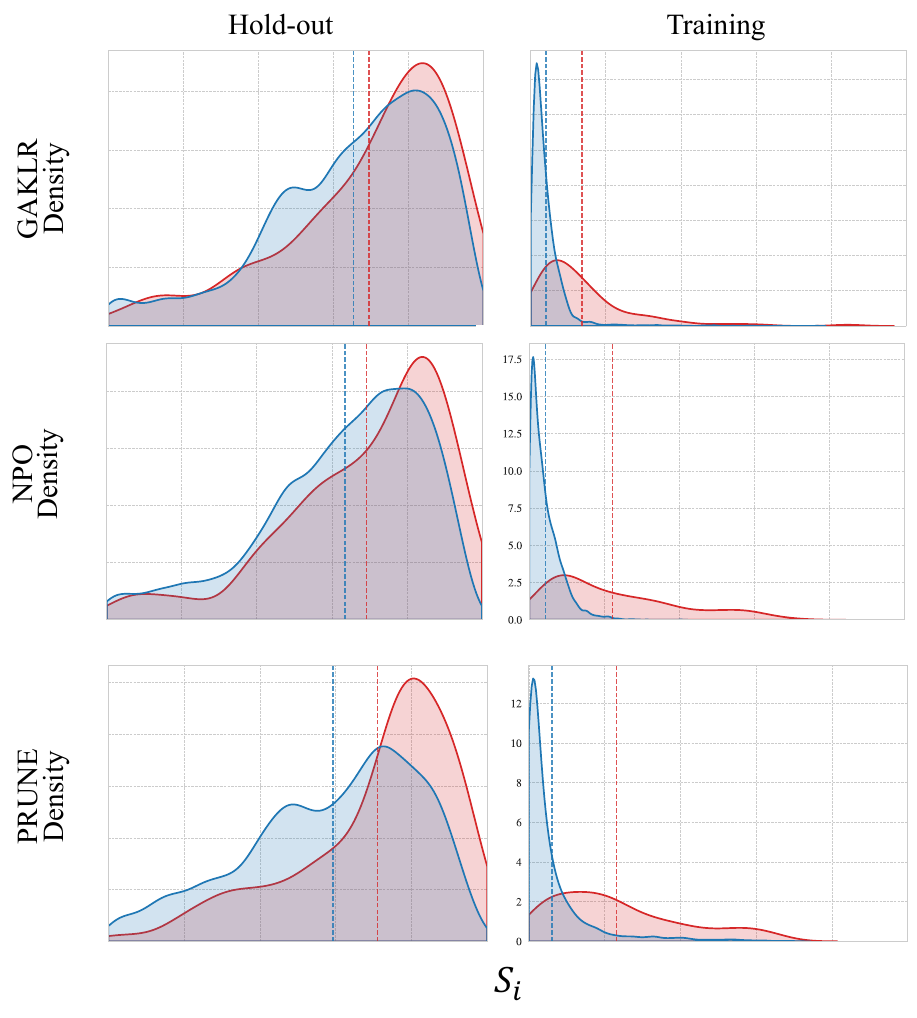}
        \label{fig:p3_prob_2}
    \end{subfigure}
    \caption{Distribution of probability-based $S_i$ scores for $p3$. The plots illustrate the distributions for Hold-out Language (Hold-out), and Training Language (Training).}
    \label{fig:three_figures_p3_all_prob}

\end{figure*}

\begin{figure*}[p]
    \centering
    \begin{subfigure}[b]{1.0\textwidth}
        \centering
        \includegraphics[width=\textwidth, height=0.9\textheight, keepaspectratio]{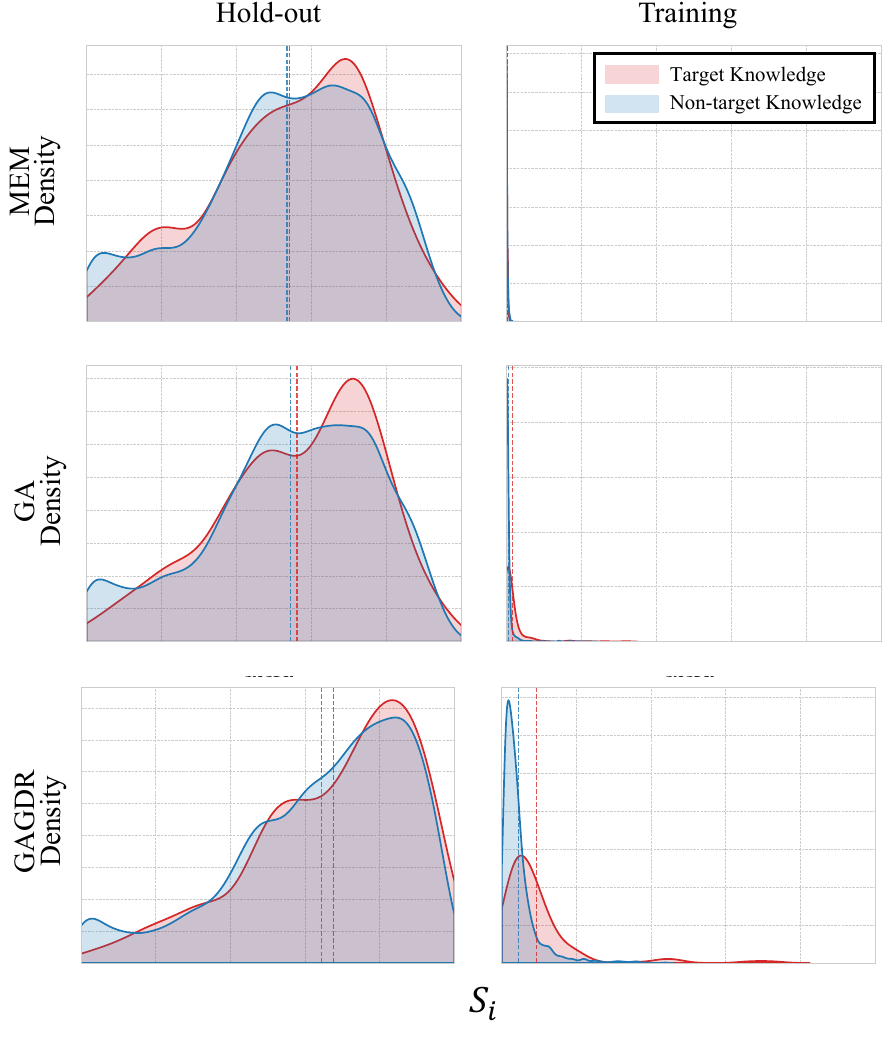}
        \label{fig:p5_prob_1}
    \end{subfigure}
\end{figure*}

\clearpage

\begin{figure*}[p]
    \ContinuedFloat
    \centering
    \begin{subfigure}[b]{1.0\textwidth}
        \centering
        \includegraphics[width=\textwidth, height=0.9\textheight, keepaspectratio]{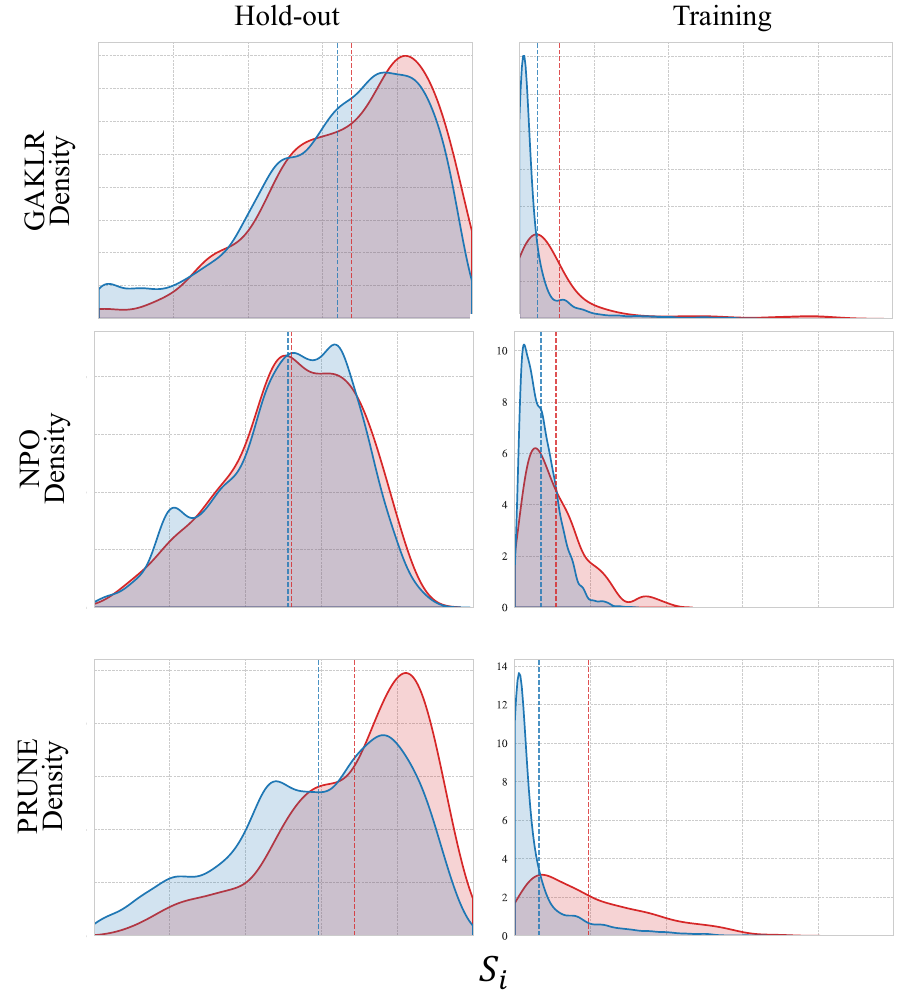}
        \label{fig:p5_prob_2}
    \end{subfigure}
    \caption{Distribution of probability-based $S_i$ scores for $p5$. The plots illustrate the distributions for Hold-out Language (Hold-out), and Training Language (Training).}
    \label{fig:three_figures_p5_all_prob}
\end{figure*}



%

\end{document}